\documentclass[journal]{IEEEtran}
\ifCLASSINFOpdf
\else
\fi
\usepackage{amsmath}
\usepackage{array}

\usepackage{stfloats}
\usepackage{url}

\usepackage{booktabs}       
\usepackage{amsfonts}       
\usepackage{caption}
\usepackage{xcolor}

\usepackage[colorlinks]{hyperref}    
\usepackage{url}            
\usepackage{nicefrac}       
\usepackage{microtype} 
\usepackage{lipsum,amsmath}
\usepackage{graphicx}
\usepackage[colorinlistoftodos]{todonotes}
\usepackage{amssymb,amsmath,amsthm}
\newtheorem{theorem}{Theorem}
\usepackage{pifont}
\newcommand{\cmark}{\ding{51}}%
\newcommand{\xmark}{\ding{55}}%

\newcommand{\vf}{{\boldsymbol{f}}}
\newcommand{\vx}{{\mathbf{x}}}
\newcommand{\vw}{{\mathbf{w}}}
\newcommand{\vk}{{\mathbf{k}}}
\newcommand{\vy}{{\mathbf{y}}}
\newcommand{\vu}{{\mathbf{u}}}

\newcommand{\vzero}{{\mathbf{0}}}
\newcommand{\vs}{{\mathbf{s}}}

\newcommand{\bbR}{{\mathbb{R}}}

\newcommand{\bbE}{{\mathbb{E}}}
\newcommand{\gp}{{\mathcal{GP}}}
\newcommand{\N}{{\mathcal{N}}}
\newcommand{\D}{{\mathcal{D}}}
\newcommand{\loss}{{\mathcal{L}}}
\newcommand{\complexity}{{\mathcal{O}}}
\newcommand{\F}{{\mathcal{F}}}

\newcommand{\vepsilon}{\boldsymbol{\epsilon}}

\newcommand{\freq}[1]{\hat{#1}} 

\newcommand{\varn}{{{\sigma}^2_{n}}}
\newcommand{\Var}{{\Sigma}}

\newcommand{\tra}{^{\top}}
\newcommand{\vmu}{{\boldsymbol{\mu}}}

\newcommand{\gprn}{\textit{SM-LMC}}
\newcommand{\csm}{\textit{CSM}}

\newcommand{\mtgp}{\textit{MTGP}}
\newcommand{\sm}{\textit{SM}}
\newcommand{\deep}{\textit{Deep}}

\newcommand{\nsm}{\textit{NSM}}

\newcommand{\nn}{\textit{NN}}

\newcommand{\mm}{m,m'}

\newcommand{\gibbs}{\textit{Gibbs}}

\begin{document}

\title{Recent Advances in Data-Driven Wireless Communication Using Gaussian Processes: A Comprehensive Survey}

\author{Kai Chen~\IEEEmembership{},
    Qinglei Kong~\IEEEmembership{},
    Yijue Dai~\IEEEmembership{},
    Yue Xu~\IEEEmembership{},
    Feng Yin~\IEEEmembership{Senior Member},
    Lexi Xu~\IEEEmembership{},
    and~Shuguang Cui~\IEEEmembership{IEEE Fellow}
    
    \thanks{Kai Chen is with Future Network of Intelligence Institute (FNii), The Chinese University of Hong Kong, Shenzhen 518172, China, with School of Information Science and Technology, University of Science and Technology of China, Hefei 230026, China (e-mail: chenkai@cuhk.edu.cn).}
    \thanks{Qinglei Kong is Future Network of Intelligence Institute (FNii), The Chinese University of Hong Kong, Shenzhen 518172, China, with School of Information Science and Technology, University of Science and Technology of China, Hefei 230026, China (e-mail: kongqinglei@cuhk.edu.cn).}
    \thanks{Yijue Dai is with Future Network of Intelligence Institute (FNii), The Chinese University of Hong Kong, Shenzhen 518172, China (e-mail: yijuedai@link.cuhk.edu.cn).}
    \thanks{Yue Xu is with Alibaba Group, Hangzhou 310052, China (e-mail: xuy.bupt@qq.com).}
    \thanks{Feng Yin is with Future Network of Intelligence Institute (FNii), The Chinese University of Hong Kong, Shenzhen 518172, China, with Shenzhen Research Institute of Big Data, The Chinese University of Hong Kong, Shenzhen 518172, China (e-mail: yinfeng@cuhk.edu.cn).}
    \thanks{Lexi Xu is with Research Institute, China United Network Communications Corporation, Beijing 100048, China (e-mail: davidlexi@hotmail.com).}
    \thanks{Shuguang Cui is with Future Network of Intelligence Institute (FNii), The Chinese University of Hong Kong, Shenzhen 518172, China, with Shenzhen Research Institute of Big Data, The Chinese University of Hong Kong, Shenzhen 518172, China (e-mail: shuguangcui@cuhk.edu.cn).}}

%
%

\markboth{Journal of \LaTeX\ Class Files,~Vol.~14, No.~8, August~2015}%
{Shell \MakeLowercase{\textit{et al.}}: Bare Demo of IEEEtran.cls for IEEE Journals}
%



\maketitle



\begin{abstract}
    Data-driven paradigms are well-known and salient demands of future wireless communication.
    Empowered by big data and machine learning, next-generation data-driven communication systems will be intelligent with the characteristics of expressiveness, scalability, interpretability, and especially uncertainty modeling, which can confidently involve diversified latent demands and personalized services in the foreseeable future. 
    In this paper, we review a promising family of nonparametric Bayesian machine learning methods, i.e., Gaussian processes (GPs), and their applications in wireless communication.
    Since GPs achieve the expressive and interpretable learning ability with uncertainty, it is particularly suitable for wireless communication. Moreover, it provides a natural framework for collaborating data and empirical models (DEM).
    Specifically, we first envision three-level motivations of data-driven wireless communication using GPs. Then, we present the background of the GPs in terms of covariance structure and model inference. The expressiveness of the GP model using various interpretable kernel designs is surveyed, namely, stationary, non-stationary, deep, and multi-task kernels. Furthermore, we review the distributed GPs with promising scalability, which is suitable for applications in wireless networks with a large number of distributed edge devices. Finally, we list representative solutions and promising techniques that adopt GPs in wireless communication systems.
\end{abstract}

\begin{IEEEkeywords}
wireless communication; Gaussian process; machine learning; kernel; interpretability; uncertainty
\end{IEEEkeywords}

\section{Introduction}

{R}ecently, there has experienced an explosion of works 
in artificial intelligence (AI) for wireless communications \cite{DBLP:journals/jsac/XuYXLC19,DBLP:conf/icc/XuYXLC19,DBLP:journals/cm/XuYXLLC20,liu2019machine,chowdhury20206g,yan2018big,hu2020cooperative}.
Furthermore, traditional paradigms (TPs) based on mathematical modeling have greatly hindered the progress of future wireless communications and negatively affected its emerging applications, such as the Internet of Vehicles (IoV) \cite{zhou2020evolutionary}, Internet of Things (IoT) \cite{samie2019cloud,casolla2019exploring}, augmented/virtual reality (AR/VR) \cite{qiao2019web,qiao2019mobile}, and energy efficient 5G \cite{zhang2019receding,hua2019energy}.
Increasingly, many new breeds of smart connected sensors and AI-enabled applications heavily depend on intelligent real-time response and explainable decision making, e.g., emergency braking in self-driving vehicles, obstruction warning for drones, fault diagnosis for intelligent manufacturing, environmental perception for cooperative multirobot systems, and predictable human-computer interaction for AR/VR, to reduce response times and human-interventions. These application-driven requirements demand the next-generation communication systems to be intelligent with the following welcome features: flexibility, scalability, interpretability, and especially uncertainty modeling to confidently involve latent demands and personalized service in the future.

Compared with traditional paradigms in wireless communication, a significant advantage of machine learning is its capability of gaining knowledge and automatically extracting information without specific rules \cite{bishop2006PRML}.
However, due to the insufficient interpretation of state prediction, 
machine learning methods with black-box decision making \cite{MacKay1998IntroductionTG,Rasmussen2006,neal2012bayesian} always confuse the diagnosis and analysis of complex communication systems and lead to a passive understanding of its functioning mechanism. 
To promote interpretable machine learning for data-driven wireless communications, in this paper, we review the Gaussian process (GP) model, and present their applications in wireless communications due to their interpretable learning ability with uncertainty.

GP is a generalization of the Gaussian probability distribution, which means GP is any distribution over functions $f({\vx})$ such that any finite set of function values has a joint Gaussian distribution \cite{bishop2006PRML,Rasmussen2006}. The GP provides a model where a posterior distribution over the unknown function is maintained as evidence is accumulated. This allows GPs to learn the underlying functions of wireless communication systems when a large number of observations are collected.
In contrast to the popular deep neural network (DNN) \cite{lecun2015deep} and other learning models, GP model shows a unique property of uncertainty qualification with a closed-form mathematical expression of great value to data-driven wireless systems that demand controllable and understandable state prediction.  
In table \ref{tab:hyp}, we compare GPs with DNN, reinforcement learning (RL), and TPs in terms of model expressiveness, interpretability, scalability, uncertainty modeling, sample efficiency, and collaboration of data and empirical models (DEM).

\subsection{Related Work}

A GP can model a large and complex wireless communication system through the design of its covariance function (also called kernel function), which encodes one's assumption about the auto-covariance of an unknown function. Therefore, a kernel is crucial in a GP model, as it implies the characteristics of distribution over functions. In addition, scalable inference is another core aspect in GP because the computational complexity of GP is cubic $\complexity(n^{3})$ for large scale wireless communication systems. This usually prevents the GP from learning a big data problem. Thus, the most important advances in wireless communication using GPs are related to both the kernel function design and scalable inference, which are extensively studied
\cite{DBLP:journals/cm/XuYXLLC20,DBLP:journals/jsac/XuYXLC19,DBLP:journals/jstsp/YinG17,xu2019scalable}.

Generally, for kernel function designation, there are broadly four categories of covariance design in the existing works for GP, including (1) compositional kernel design \cite{rios2019compositionally,chen2017hierarchically}, where kernels are constructed compositionally from several existing base kernels; (2) spectral kernel learning, where kernels are derived by modeling the kernel spectral density as a mixture of distributions \cite{Wilson2013,chen2019incorporating,remes2017non,DBLP:journals/tsp/YinPCTLZ20}; (3) deep kernel representation \cite{wilson2016deep,pmlr-v124-dai20a}, where DNN plays a role in nonlinear mapping between input space and feature space; and (4) multi-task kernel \cite{bonilla2008multi,Ruan2017}, where adjacent devices (tasks) share knowledge and interact with each other to obtain collective intelligence.
In the next sections, we review related works in detail.

To overcome the computational complexity issue of GP \cite{DBLP:journals/cm/XuYXLLC20,Rasmussen2006,DBLP:journals/jsac/XuYXLC19,Rasmussen2010} used in large scale wireless communication systems, scalable inference can be achieved by exploring (1) low-rank covariance matrix approximation \cite{williams2001using,williams2006gaussian},  (2) special structures of the kernel matrix \cite{wilson2015kissgp,saatcci2011scalable},
(3) Bayesian committee machine (BCM), which distributes computations to a big number of computing units \cite{Tresp00,Quinonero-Candela2005}, (4) variational Bayesian
inference \cite{hensman2015scalable,hensman2018variational}, and (5) special optimization \cite{DBLP:conf/aaai/LingLJ16,DBLP:journals/tsp/YinPCTLZ20}. Notably, these scalable methods are not exclusive, and we can combine some of them to get a better method, for instance, stochastic variational inference (SVI) \cite{hensman2015scalable,hensman2018variational} combining the strength of inducing points for low-rank approximation and variational inference.

\subsection{Outline}
The main contributions of this survey are summarized below:
\begin{itemize}
    \item We extensively discuss the generally desired AI features of next-generation data-driven wireless communication systems, namely, expressiveness, scalability, interpretability, and uncertainty modeling. Regarding these aspects, we compare GP with other machine learning methods and then conclude that GP can cover these qualities better (as shown in Table \ref{tab:hyp}).
    
    \item We broadly analyze and explain four categories of covariance design in terms of mathematical theorem and GP kernel expression, including (1) stationary kernel, (2) non-stationary kernel, (3) deep kernel, and (4) multi-task kernel. These kernels leverage both the expressiveness and interpretability of the GP model in wireless communication.
    
    \item Due to the scalability demand and distributed deployment of wireless communication systems, we review and evaluate the advances of distributed GP with scalable inference for big data of cloud intelligence as well as AI-enabled edge devices.
    
    \item We show an exemplary case by extrapolating the number of online 5G users collected from a real world 5G wireless base station.
    The results show the expressiveness, interpretability, and uncertainty modeling of GPs for data-driven wireless communication.
    
    \item We exhibit some representative wireless communication scenarios for applying the GP models and further envision some open issues and challenges of using GPs for future data-driven wireless communication.
\end{itemize}

For the rest of this paper, we begin by introducing the motivation of using GPs for data-driven wireless communication in section \ref{sec:motivation} and then give the mathematical background of GPs in section \ref{sec:gp}. In section \ref{sec:advance} and section \ref{sec:dis-gp}, we present the advances of GPs.
A demonstration for wireless communication using GPs is given in section \ref{sec:gp-example}.
In section \ref{sec:wireless-gp} and section \ref{sec:future}, we give existing GP applications and future research on wireless communications, respectively.

\section{Data-driven wireless communication: Motivation with unique features}\label{sec:motivation}

\begin{table*}[h!]
    \scriptsize
    \begin{center}
        \begin{tabular}{ccccccc}
            \toprule
            {Method} & {Expressiveness} & {Interpretability} & {Scalability} & {Uncertainty modeling} & {Sample efficiency} & {Collaboration of DEM} \\ 
            \midrule
            TPs \cite{ghazal2016non,he2018mobility} & \xmark & \cmark & \xmark & \xmark & \cmark & \xmark \\
            DNN \cite{lecun2015deep,zappone2019wireless} &  \cmark & \xmark & \cmark & \xmark & \xmark & \xmark \\
            RL \cite{hu2020cooperative,ning2020joint} & \cmark & \xmark & \cmark & \cmark & \xmark & \xmark \\
            GPs \cite{Rasmussen2006,xu2019wireless} & \cmark & \cmark & \cmark & \cmark & \cmark & \cmark \\
            \bottomrule
        \end{tabular}
    \end{center}
    \caption{Comparisons between GP and other popular methods in terms of AI characteristics for wireless communication.}
    \label{tab:hyp}
\end{table*}

In this section, we present the unique features for the next-generation data-driven wireless communication using machine learning methods with expressiveness, scalability, uncertainty modeling, and interpretability (see Table \ref{tab:hyp}).

Due to the inherent intelligence requirements in data-driven wireless communication systems,
there are three levels of motivations to apply GPs.
First, the low-level motivation is based on the demands of smart, efficient, and flexible decision making, planning, and prediction in future wireless communication systems \cite{liu2019machine}, which cannot be achieved by applying traditional paradigms. Then, the comparison between GP and other machine learning methods brings the middle-level motivation and comprehensively explains why we tend to choose the GP model 
for data-driven wireless communication systems \cite{bishop2006PRML,Rasmussen2006}. As shown in Section \ref{sec:gp-example}, the high-level motivation is derived from the competitive applications empowered by GPs in wireless communication.  Specifically, the motivations can be summarized as follows:

\begin{itemize}
    \item For future wireless communication systems, it is expected that there are many latent demands and personalized services driven by diversified applications. These latent demands and personalized services can be further modeled and improved by using machine learning methods, with the growth of historical data, and ever-increasing computing power. There are many features describing future wireless communication: (a) expressiveness correlated to model complexity which 
    results from diversified application scenarios \cite{chowdhury20206g,DBLP:journals/iotj/ChengFYC17}; (b) scalability on big data due to the ever-growing network size with network densification and an increasing number of connected intelligent devices \cite{liu2019machine,zhang2018synergy}; (c) uncertainty resulting from a dynamical communication environment \cite{liu2018covert,su2019robust}; and (d) interpretable knowledge discovery and representation for understanding the mechanism of complex systems \cite{yang20196g,chang2018learn}. In particular, uncertainty modeling is critical for state prediction in wireless networks since there are always multiple noises and dynamic factors intervening the status of the system and the mobile users' experience.

    \item As a class of Bayesian nonparametric model, GP provides a principled, practical, probabilistic approach for learning the patterns encoded by kernel structure \cite{Rasmussen2006}. Among all machine learning models, the GP has a tight connection with various learning models \cite{bishop2006PRML,MacKay1998IntroductionTG,Rasmussen2006}, including spline models, support vector machines (SVMs), regularized least-squares models, relevance vector machines (RVMs), autoregressive moving averages (ARMAs), and deep neural networks (DNNs). In particular, GPs have advantages with respect to the interpretation of model learning, model selection, and uncertainty prediction from Bayesian point of view. Using an appropriate kernel structure and computational approximation, GP can model any function of wireless communication systems with flexibility and scalability. Owing to the Bayesian rules, GP with a measure of uncertainty is more robust to overfitting problems.
    In comparison with other machine learning models, 
    the GP model can simultaneously meet the requirements of expressiveness, scalability, uncertainty modeling, and interpretability \cite{bishop2006PRML,Rasmussen2006} for data-driven wireless communications.

    \item Thanks to the Bayesian properties, GP model has eye-catching interpretations in terms of model construction, selection, and hyper-parameter adaptation (see section \ref{sec:gp}). Such interpretation strengths promote a large number of GP models to empower diversified wireless communication applications. 
    There are five popular GP models using different kernels to support various wireless communication tasks, such as the GP models with stationary spectral mixture (SM) \cite{Wilson2013,zou2017adaptive} and compositional kernels \cite{Duvenaud2013} (see section \ref{sec:stationGP}), non-stationary (NS) kernels \cite{DBLP:journals/jsac/XuYXLC19,Remes2017,shen2019learning,wang2018deepmap,jung2013compressive} (see section \ref{sec:nsgp}), deep kernels \cite{deepkernel,teng2018localization} (see section \ref{sec:dkgp}), and multi-task kernels \cite{DBLP:journals/jsac/XuYXLC19,bonilla2008multi} (see section \ref{sec:mtgp}). 
    Furthermore, GPs have scalability variations with distributed inference to scale large data on a big number of edge devices (see section \ref{sec:dis-gp}).
    The distributed GPs can make full use of the computational resources of local edge devices in wireless networks to gain efficiency improvement as well as privacy protection \cite{yin2017distributed,salari2017distributed}.
\end{itemize}

\section{Background of Gaussian process for machine learning}\label{sec:gp}

There are multiple uncertainty issues in the modeling of wireless communication: (1) functional uncertainty describing the gap between the true function and learned model ; (2) prediction uncertainty with a fuzzy range caused by the amount of observed evidence; (3) input uncertainty due to the noise generated during the wireless propagation; and (4) output uncertainty due to unstable wireless propagation and poor precision of measuring sensors. Theoretically, these uncertainties, as well as interpretability, can be well represented by a GP model.
In this section, we briefly describe the background of Gaussian process for machine learning in terms of its mathematical definition, kernel function and model inference.

\begin{figure*}[h!]
    \centering
    \renewcommand{\tabcolsep}{0.5mm}
    \begin{tabular}{p{0mm}*{4}{c}}
        & \includegraphics[width=0.5\columnwidth]{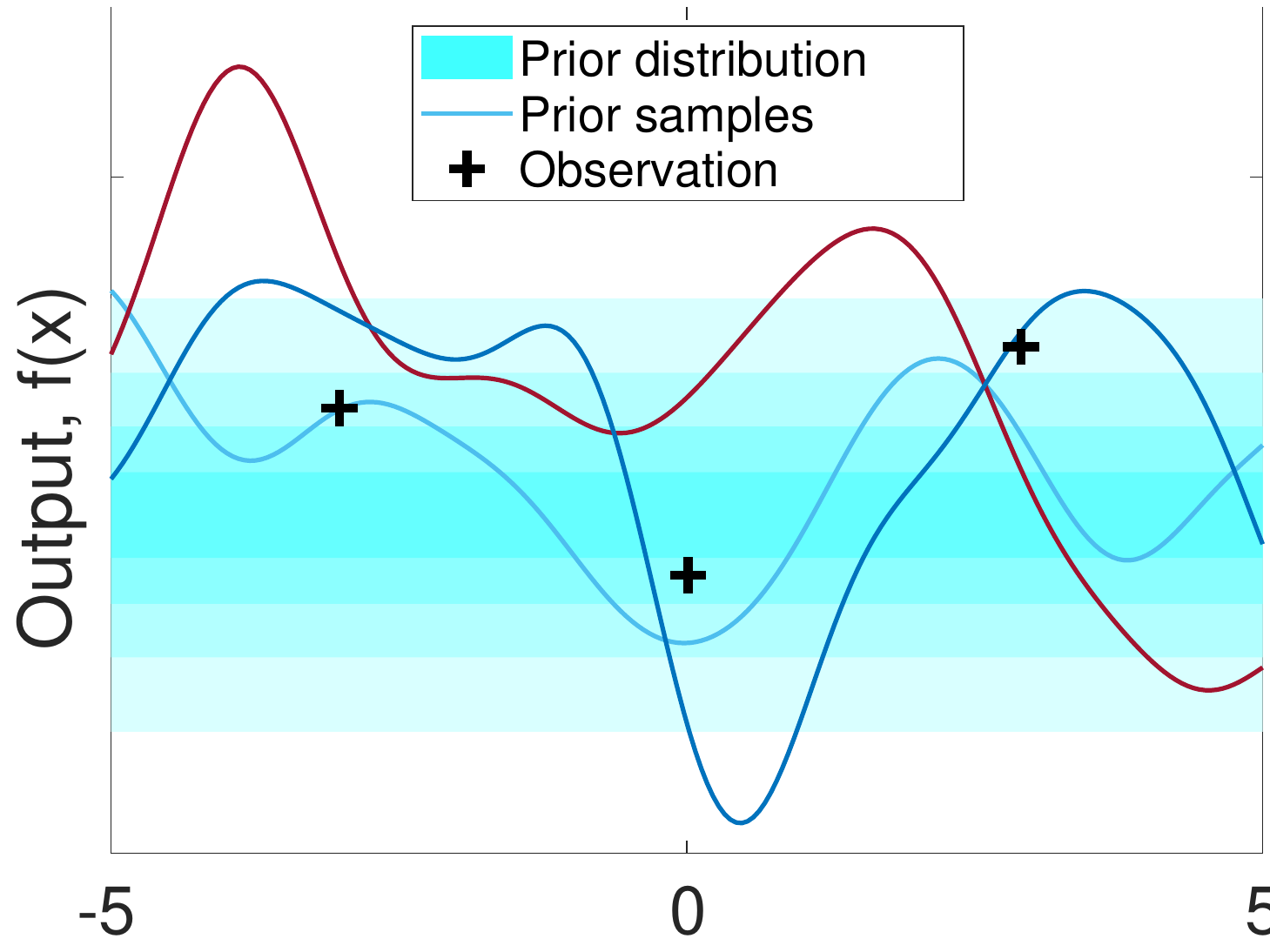} 
        & \includegraphics[width=0.5\columnwidth]{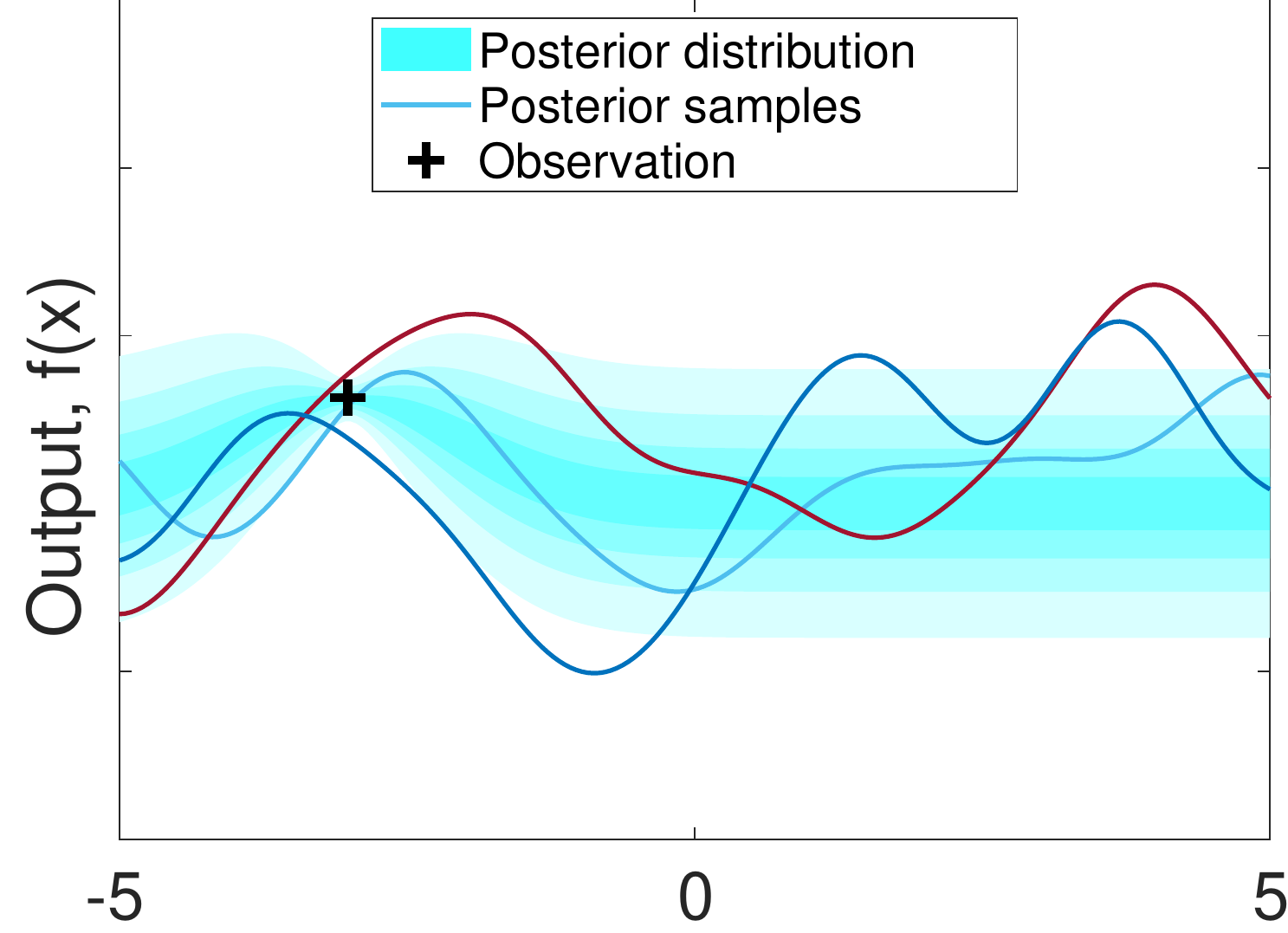}
        & \includegraphics[width=0.5\columnwidth]{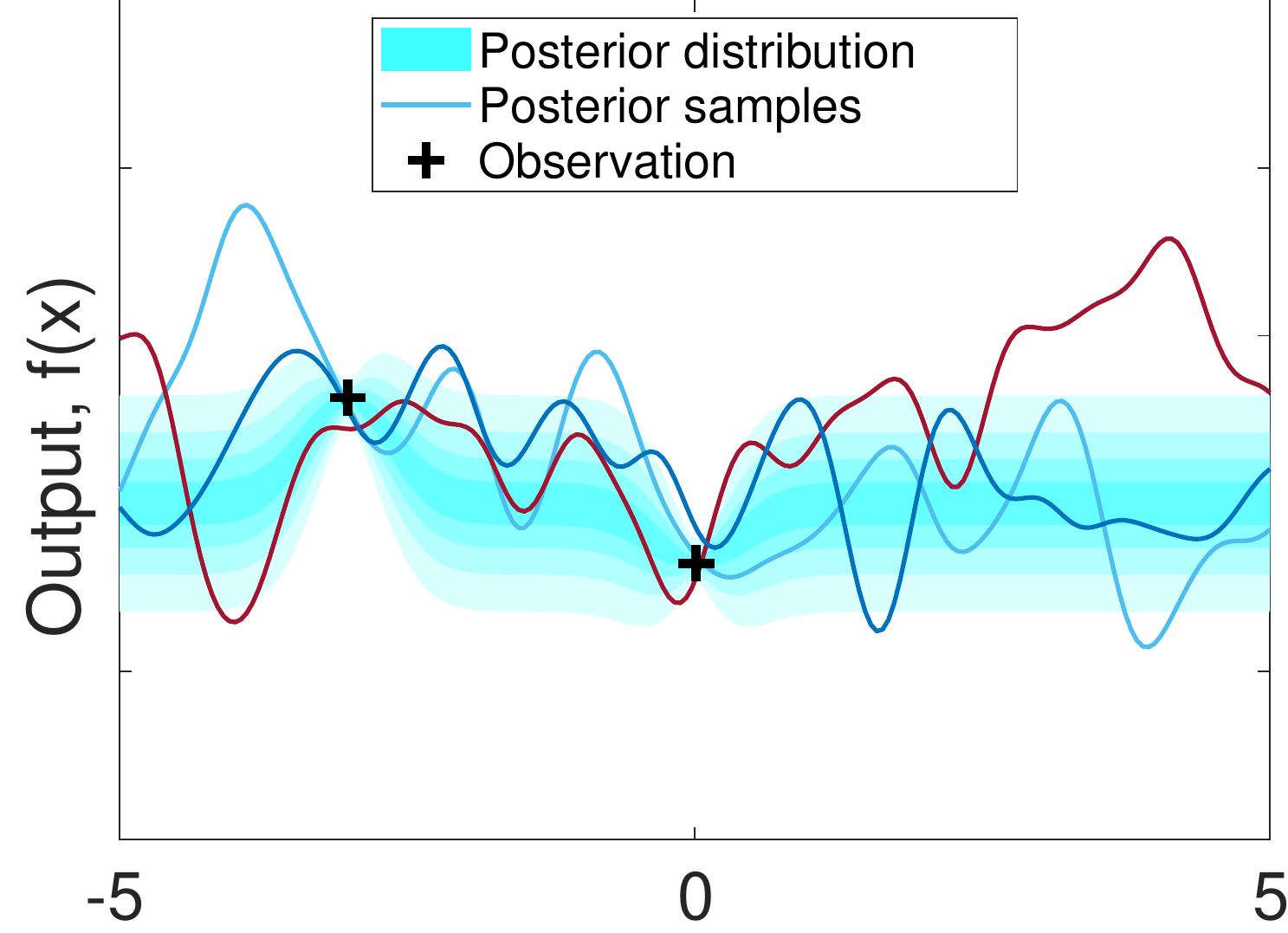}
        & \includegraphics[width=0.5\columnwidth]{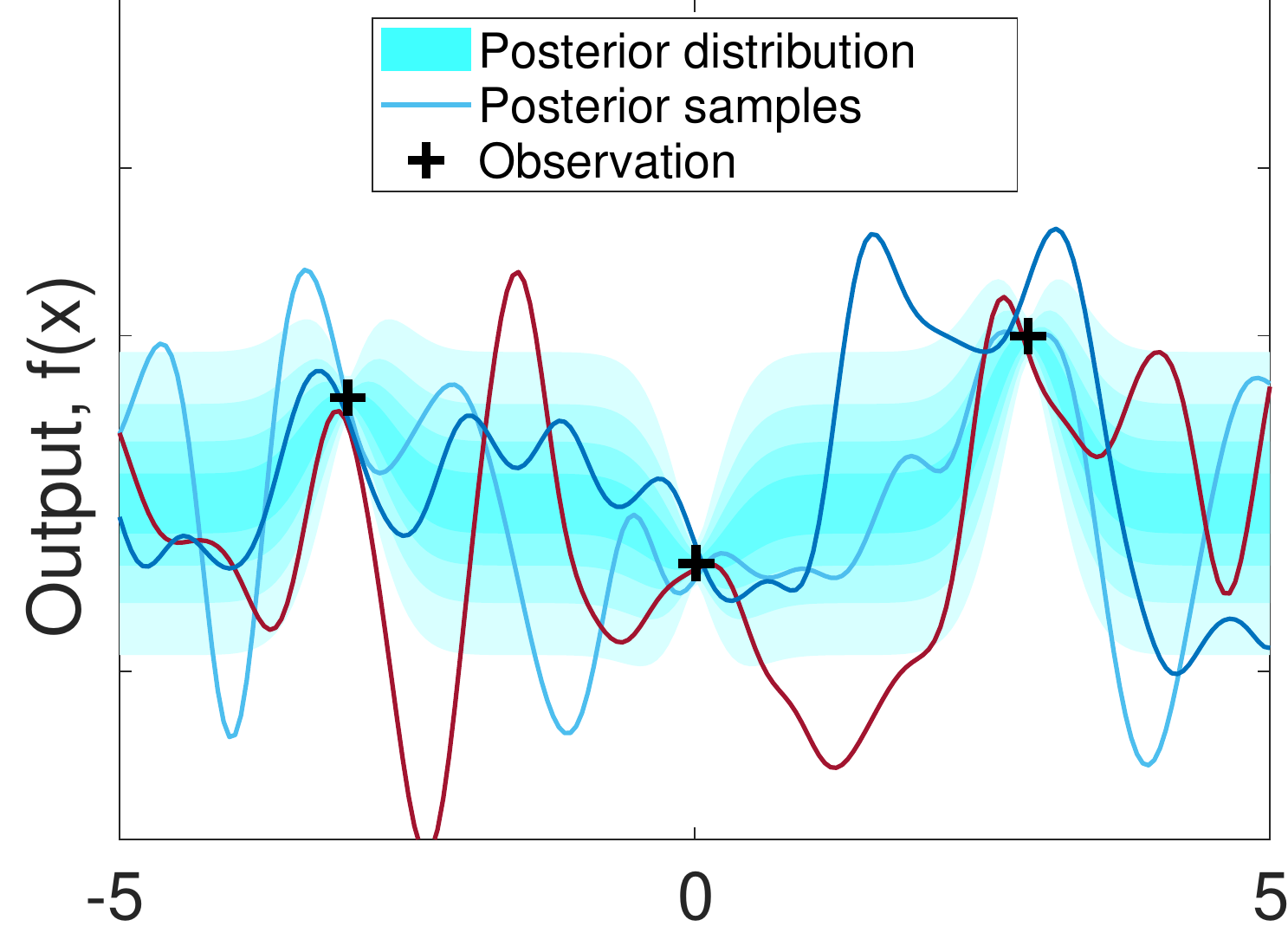}\\
        & (a)   & (b)  & (c) &  (d)
    \end{tabular}
    \caption{Samples from GP prior distribution and GP posterior distribution based on three observations (black crosses). Subplot (a) is the prior distribution (in cyan) and sampling (in light blue, dark blue, and red); subplots (b), (c), and (d) are the posterior distribution and sampling with one, two, and three observations, respectively. The shaded area (in cyan) can be seen as the uncertainty bound of the predictive function value. With the increase in collected observations, GPs can adapt the underlying function space very smoothly.}
    \label{fig:gp-fun}
\end{figure*}

\subsection{Definition of Gaussian process }\label{sec:def-gp}

From the function-space view, a Gaussian process \cite{MacKay1998IntroductionTG,Rasmussen2006} defines a distribution $p(f(\vx_{1}), f(\vx_{2}), ..., f(\vx_{n}))\sim \N(m(\vx), K(\vx, \vx'))$ over functions, completely specified by its first and second-order statistics, namely, the mean $m({\vx})$ and the covariance $k({\vx}, {{\vx}{'}})$ functions \cite{DBLP:books/lib/Bishop07}. For a given input location ${\vx}\in{\bbR}^{p}$ of a real stochastic process $f(\vx)$, the mean $m({\vx})$ and covariance function $k({\vx}, {{\vx}{'}})$ are defined as:
\begin{subequations}
    \begin{align}
    m({\vx})&=\bbE\left[f(\vx)\right]\\
    k({\vx}, {\vx'})&=\bbE[(f(\vx)-m(\vx))(f(\vx')-m(\vx'))]
    \end{align}
\end{subequations}
Thus, a GP is expressed as $f({\vx})\sim{\gp}(m({\vx}), k({\vx}, {{\vx}{'}}))$.
Without loss of generality, the mean of a GP often assumed to be zero anywhere because we usually do not have any
prior knowledge about the mean. The covariance function (also called the kernel) between function values is applied to construct a positive definite covariance matrix on input points $X$ for the joint Gaussian distribution, here denoted by Gram matrix $K=K({X}, {X})$. By using a GP prior over functions in the kernel designation and parameter initialization, from the training data $X$, we can predict the unknown function value $\tilde{y}_{*}$ and its variance $\mathbb{V}[{y_{*}}]$ (that is, its uncertainty) for a test point ${\vx}_{*}$. Specifically, we have the following predictive equations for GP regression \cite{Rasmussen2006,Rasmussen2010}:
\begin{subequations}
    \begin{align}
    \tilde{y}_{*}&=\vk_{*}\tra(K+{{\varn}}I)^{-1}{\vy}
    \label{eq:pred-mean}\\
    \mathbb{V}[\tilde{y_{*}}]&=k({\vx}_{*}, {\vx}_{*})-\vk_{*}\tra(K+{{\varn}}I)^{-1}\vk_{*},
    \label{eq:pred-var}
    \end{align}
\end{subequations}
where $\vk_{*}\tra$ is the covariance vector between ${\vx}_{*}$ and $X$, $\varn$ is the variance of the noise, and ${\vy}$ is the vector of observations corresponding to $X$.

\subsection{Gaussian process kernel}\label{sec:cov-gp}

Basically, the smoothness and generalization properties of GP depend on the kernel function and its hyper-parameters ${\Theta}$. Choosing an appropriate kernel function and the corresponding initial hyper-parameters are crucial to GP design since the posterior distribution can vary significantly for different kernels.
The most extensively used covariance function is stationary. We introduce a generalized theory of both stationary and non-stationary covariance functions in the later sections. For the underlying function to be modeled by the Gaussian process, there are many characteristics, such as exponentially decayed dependency and periodic dependency, which can be encoded by specific covariance functions.

To make the GP model applicable for practical applications, the inference of the GP model is also very important. During the inference phase of the GP model, the freedom of model selection is considerable even though an appropriate covariance was specified in advance. Typically, GPs contain hyper-parameters ${\Theta}$ describing the properties of the kernel and noise of the GP. Suppose we have chosen a covariance function $k(\vx, \vx')$ with hyper-parameters $\Theta_{k}$. The inference of the GP means Bayesian model selection with the possible best values of $\Theta=\{\Theta_{k}, \varn\}$. Such selection can be accomplished by minimizing the negative log marginal likelihood (NLML), $\loss=-\log\ p({\vy}|X,{\Theta})$.
The inference and posterior sampling of a GP model are illustrated in Fig. \ref{fig:gp-fun}.

The NLML can be used for assessing the goodness of fit of the GP model. 
For the evaluation of GP model, we usually apply the mean squared error (MSE) and mean absolute error (MAE) to measure prediction performance. Specifically, the predictive uncertainty described in Eq. (\ref{eq:pred-var}) scores the confidence of the prediction.

\section{Advances in GP kernels}\label{sec:advance}
\subsection{Stationary spectral mixture kernel}\label{sec:stationGP}

Data generated in wireless communication systems often demonstrate the following patterns: (1) weekly periodic trends on weekdays and weekends, (2) daily periodic trends in working hours and spare time, (3) decayed deviations in terms of small-scale variation, and (4) some noise introducing disorder fluctuations. These patterns are generally stationary and can be captured by the GP with a flexible kernel structure (see section \ref{sec:gp-example}). However, without tangible prior information, the number of patterns and their signal features are not clear for the definition and construction of a GP model.
Alternatively, we can apply a universal representation of stationary kernels and then automatically infer the latent patterns through optimization, which can simplify the practice of machine learning in wireless communication systems and enhance the efficiency of interpretable knowledge discovery.

In this section, we review the theoretical foundation of stationary covariance functions and recent GP works. Stationary covariance is regarded as a function of $\tau=\vx-\vx'$ other than input location $\vx$, which is invariant to translations in the input space \cite{Rasmussen2006}. For each covariance function of a stationary process, there is a corresponding representation, the Fourier transform of a positive finite measure $\psi$, in the frequency domain. Referring to \cite{Stein,Bochner2016}, Bochner's theorem indicates the connection between the covariance function and its spectral density.

\begin{figure}[h!]
    \centering
    \renewcommand{\tabcolsep}{-2.0mm}
    \begin{tabular}{p{0mm}*{2}{c}}
        & \includegraphics[width=0.55\columnwidth]{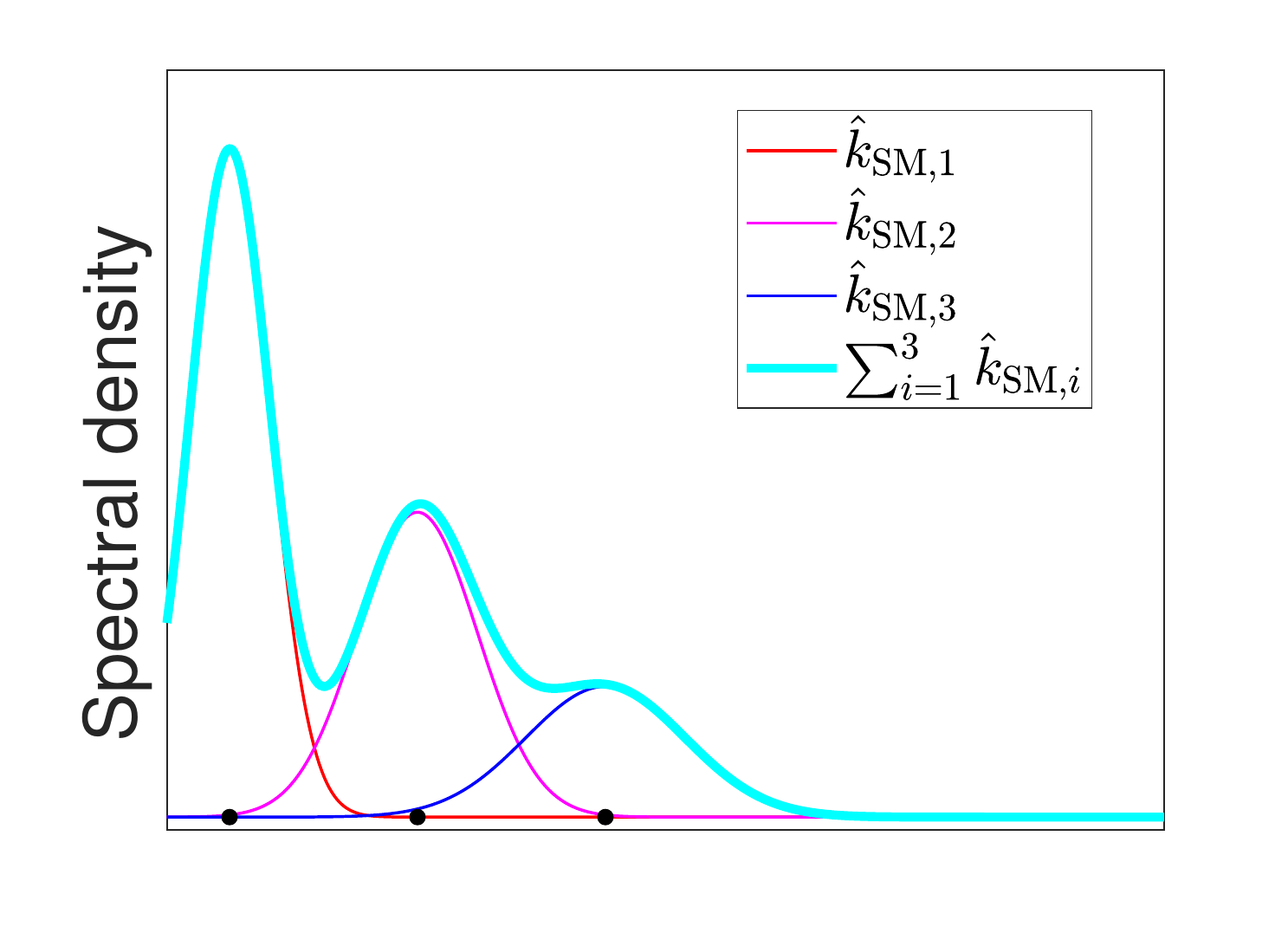}
        & \includegraphics[width=0.55\columnwidth]{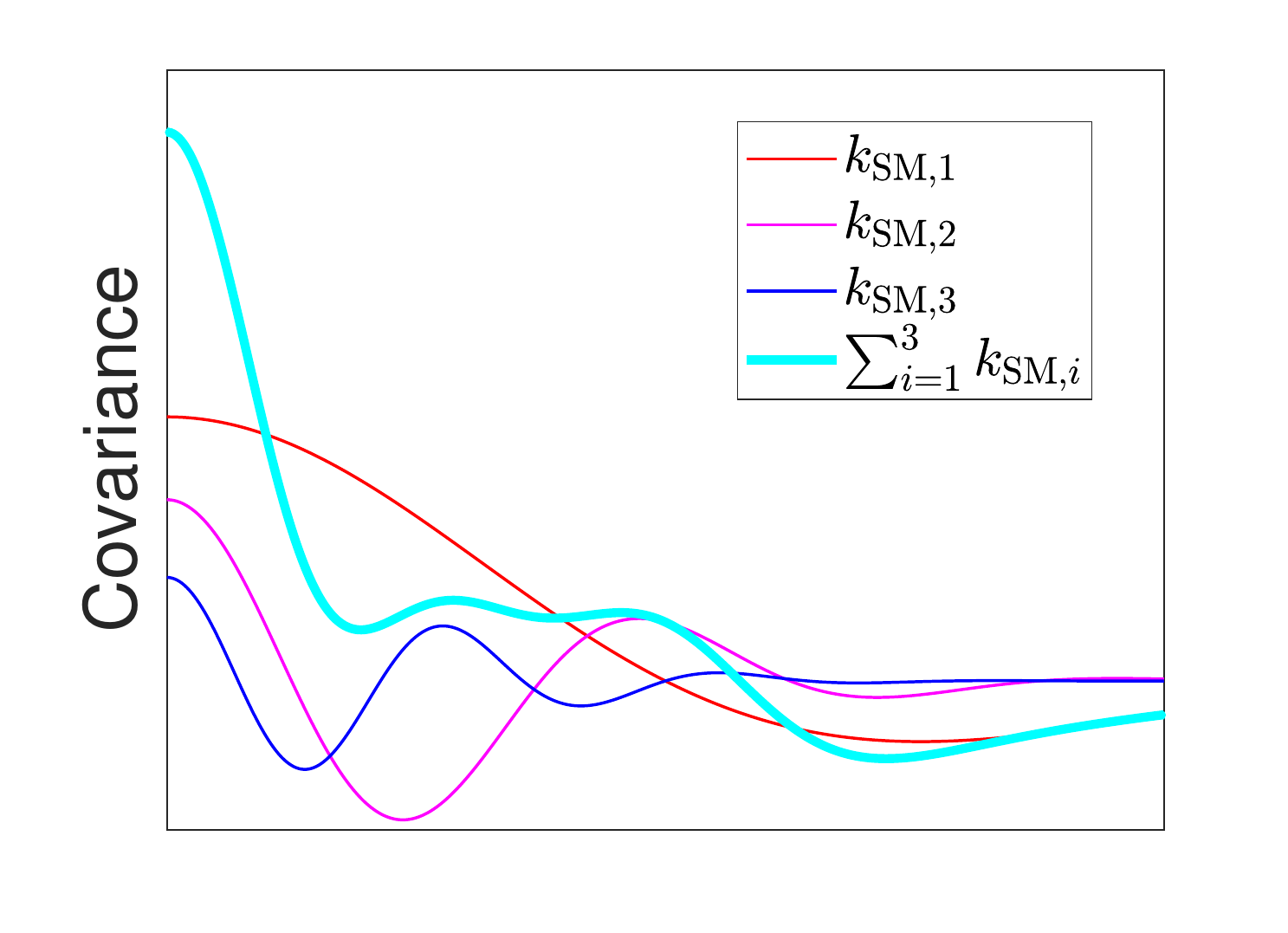}\\
        & (a) Spectral densities of SM  & (b) Covariance of SM
    \end{tabular}
    \caption{Spectral densities (left) with a mixture of Gaussians and corresponding covariance functions (right) in the SM kernel. For SM, the location (black dot) of each component denotes the period of underfunction.}
    \label{fig:sm}
\end{figure}

\begin{theorem}[Bochner's Theorem \cite{Stein,Bochner2016}]\label{th:bochner}
    \textit{A complex-valued function $k$ on $\bbR^P$ is the covariance function of a weakly stationary mean square continuous complex-valued random process on $\bbR^P$ if and only if it can be represented as
        $$k(\tau)= \int_{\bbR^P} e^{2\pi\jmath\vs \tra\tau}\psi(d\vs),$$
        where $\psi$ is a positive finite measure and $\jmath$ denotes the imaginary unit.}
\end{theorem}
If $\psi$ has a density $\freq{k}(\vs)$ called the spectral density or power spectrum of $k$, Theorem (\ref{th:bochner}) implies the following Fourier dual.
\begin{align}
\begin{cases}
k(\tau)&= \int \freq{k}(\vs) e^{2\pi \jmath\vs \tra\tau} d\vs, \\
\freq{k}(\vs)& = \int  k(\tau) e^{2\pi \jmath \vs \tra\tau} d\tau.
\end{cases}
\end{align}

Based on Bochner's theorem, a large number of expressive stationary kernels are proposed, including the spectral mixture kernels (SMs) and compositional kernels. Compositional kernels \cite{Duvenaud2011,Duvenaud2013} have advanced kernel structures constructed from a combination of normal kernels by using a series of kernel operations, such as plus, wrap, and product operations. Furthermore, one of the most representative stationary kernels is the spectral mixture kernel \cite{Wilson2013,Jang2017,chen2019incorporating}, as SM can approximate any stationary kernels with a sufficient number of components. Here, we mainly introduce the SM kernel. An SM kernel $k_\sm$ is derived by representing its spectral density (the Fourier transform of a kernel) with a Gaussian mixture model (GMM) (see Fig. \ref{fig:sm}), $\freq{k}_{\sm}(\vs) =\sum_{i=1}^{Q}\freq{k}_{{\sm}, i}({\vs})$, 
where $\freq{k}_{{\sm}, i}({\vs})=w_{i}[\varphi_{\sm,i}(\vs) + \varphi_{\sm, i}(-\vs)]/2$, $Q$ is the number of Gaussians, $w_i$ is the weight of the $i$-th Gaussian, and ${\varphi}_{{\sm}, i}({\vs})={\N}({\vs};\vmu_{i},{\Var}_{i})$ is a scale-location Gaussian with mean $\vmu_i$ and variance ${\Var}_i$. The symmetrization makes $\freq{k}_{{\sm}, i}({\vs})$ even, that is, $\freq{k}_{{\sm}, i}({\vs}) = \freq{k}_{{\sm}, i}(-{\vs})$ for all $\vs$. Then, applying the inverse Fourier transform, we can obtain the SM kernel as follows:
\begin{align}\label{eq:sm}
\begin{split}
k_{\sm}(\tau)&={\F}_{s\rightarrow \tau}^{-1}\bigg[\freq{k}_{\sm}(\vs)\bigg](\tau)\\
&= \sum_{i=1}^Q{w_i}{\cos\left(2\pi\tau\tra\vmu_{i}\right)}{\exp\left(-2\pi^2\tau{\Var}_{i}\tau^{\top}\right)},
\end{split}
\end{align}
where ${\F}_{s\rightarrow \tau}^{-1}$ denotes the inverse Fourier transform operator from the frequency domain to the time domain. For the SM kernel, we can interpret $w_{i}$, $\vmu_{i}$ 
, and ${\Var}_{i}$
as the signal variance, inverse period, and inverse length scale of the $i$-th covariance component, respectively.
In summary, the SM kernel can be seen as a generalization of existing stationary kernels. Note that the GP model with an SM kernel has been used for wireless traffic prediction \cite{xu2019wireless} and is trusted by the application of wireless communication. In section \ref{sec:gp-example} we predict the number of online 5G users by using a GP model with an SM kernel.

\subsection{GP with non-stationary kernel}\label{sec:nsgp}

In addition to stationary patterns, there are also a few complex non-stationary patterns with time-varying characteristics for wireless communication, for instance, mmWave massive MIMO channel modeling \cite{liu2018novel}, 5G wireless channel modeling \cite{wu2017general}, wireless control systems \cite{eisen2018learning}, 3D non-stationary unmanned aerial vehicle (UAV) MIMO channels \cite{zhu2018novel}, non-stationary mobile-to-mobile channels allowing for velocity and trajectory variations in mobile stations \cite{dahech2017non}, and non-stationary channel modeling for vehicle-to-vehicle communications \cite{li2016cluster}.
In contrast to the stationary kernel depending only on the distance $\tau=\vx-\vx'$, the signal
characteristics of non-stationary GP, such as frequencies, amplitudes, and spectral densities, have direct dependences on the input locations $\vx$.
The extension of Bochner's theorem (see Theorem \ref{th:bochner}) to the non-stationary domain has a generalized spectral representation on the $P\times{P}$ surface
\begin{align}
k(\vx, \vx') &= \int_{\bbR^{P}}\int_{\bbR^{P}} e^{2\pi \jmath (\vx\vs - \vx'\vs')} \vu_{\textit{S}}(d\vs, d\vs'),
\end{align}
where $\vu_{\textit{S}}$ is a positive finite measure on spectral surface $P\times{P}$.

\begin{figure}[h!]
    \centering
    \renewcommand{\tabcolsep}{-2.0mm}
    \begin{tabular}{p{0mm}*{2}{c}}
        & \includegraphics[width=0.55\columnwidth]{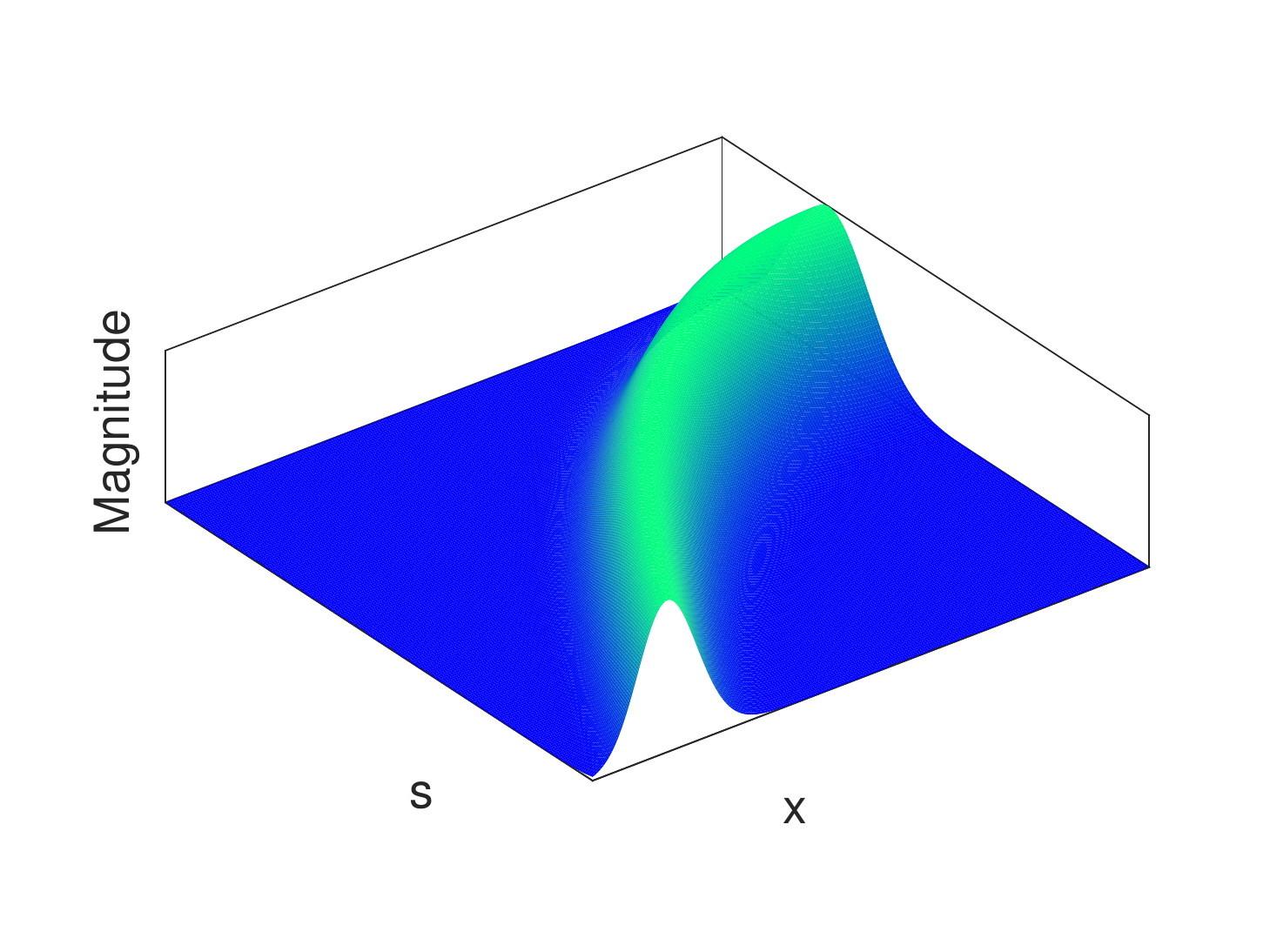}
        & \includegraphics[width=0.55\columnwidth]{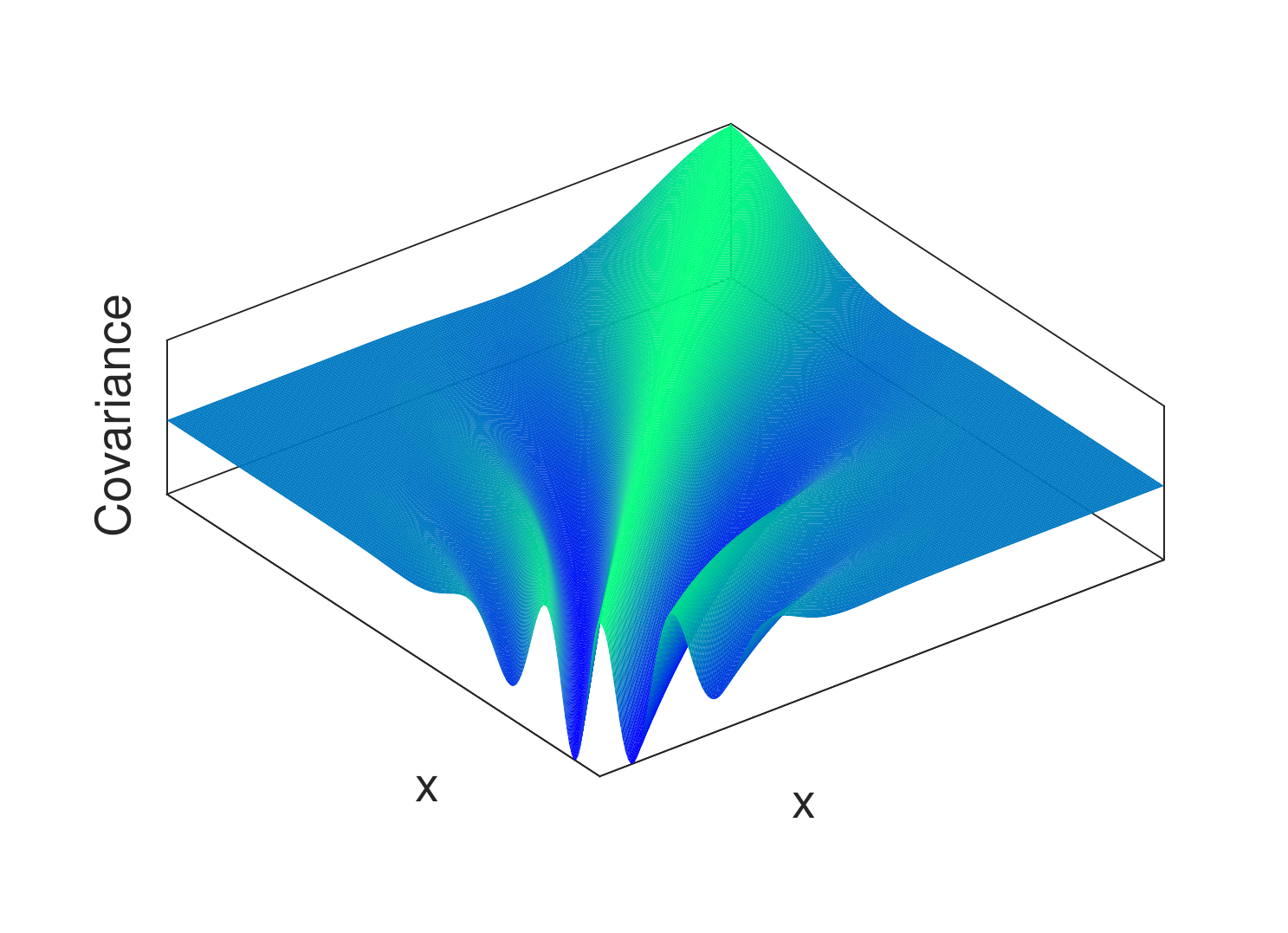}\\
        & (a) Spectrogram of NSM  & (b) Covariance of NSM
    \end{tabular}
    \caption{Spectrogram (left) depending on both input $x$ and spectral density $s$ and the corresponding covariance functions (right) for NSM. }
    \label{fig:nsm}
\end{figure}

Arguably, the dot product kernel is the simplest non-stationary kernel \cite{Rasmussen2006}. The well-known and extensively used non-stationary kernels are linear and polynomial kernels \cite{Rasmussen2006}, which are less parameterized for representing complex patterns. Since the introduction of the neural network (NN) kernel \cite{neal2012bayesian}, GPs can approximate both DNN and one hidden layer neural network model (known for universal approximator and nonlinear property) with infinity neurons.
After that, Gibbs \cite{gibbs1997bayesian} developed the non-stationary covariance function shown in Eq. (\ref{eq:gibbs}) by
considering a grid of exponential basis functions and parameterizing its length scale as positive functions,
\begin{align}\label{eq:gibbs}
\begin{split}
k_{\gibbs}(\vx,\vx')=&\prod_{p=1}^{P}\sqrt{\frac{2{\theta_{\ell, p}(\vx)}{\theta_{\ell, p}(\vx')}}{{\theta_{\ell, p}^2(\vx)}+{\theta_{\ell, p}^2(\vx')}}}\\
&\times\exp\left(-\sum_{d=1}^{D}\frac{({{x}_p}-{{x}_p'})^2}{{\theta_{\ell, p}^2(\vx)}+{\theta_{\ell, p}^2(\vx')}}\right).
\end{split}
\end{align}
Then, Higdon \cite{higdon1999non} proposed a non-stationary spatially evolving GP using a process convolution to model toxic waste remediation. Based on \cite{higdon1999non},
Paciorek \cite{DBLP:conf/nips/PaciorekS03} generalized the Gibbs kernel using non-stationary quadratic form $Q_{\vx, \vx'}=(\vx - \vx')((\Sigma_{\vx}+\Sigma_{\vx'})/2)^{-1}(\vx - \vx')$ instead of $\tau$ in any stationary kernel, where $\Sigma_{\vx}$ is the positive length scale function of input $\vx$.
After proposing the SM kernel (Eq. (\ref{eq:sm})), in \cite{Remes2017}, a non-stationary SM (NSM, see Fig. \ref{fig:nsm} ) kernel was introduced by modeling the spectral surface as a two-dimensional GMM.
\begin{align}\label{eq:nsm}
\begin{split}
k_{\nsm}(x,x') = & \sum_{i=1}^Q w_i^2 \exp(- 2 \pi^2 \tilde{\vx}^{\top} \Sigma_i \tilde{\vx})\\
&\phantom{==}\times \Psi_{\mu_i,\mu'_i}(x)^{\top}\Psi_{\mu_i,\mu'_i}(x'),
\end{split}
\end{align}
where $\tilde{\vx} = (x, -x')^{\top}$ and
$$\Psi_{\mu_i,\mu'_i}(x) = \begin{pmatrix} \cos 2\pi\mu x + \cos 2\pi\mu' x \\ \sin 2\pi\mu x + \sin 2\pi\mu' x \end{pmatrix}.$$
For the aforementioned non-stationary kernels, their hyper-parameters can be parameterized as positive functions described by stationary GPs. For example, we can parameterize ${\theta_{\ell}}$ as ${\theta_{\ell}}\sim\gp\left(0, k_{\ell}(\vx, \vx)\right)$.
Recently, the harmonizable kernel \cite{shen2019harmonizable} showed a novel spectral representation of the non-stationary kernel by incorporating a locally stationary kernel with an interpretation of the Wigner distribution function. In \cite{shen2019learning}, another convolutional spectral kernel was proposed to give a concise representation of the input frequency spectrogram, but it shows less insight into a prespecified complex-valued radial base.
To meet the development needs of a non-stationary GP, the non-separable and non-stationary kernel \cite{wang2020non}, including a varying non-separability and local structure, has a natural interpretation through the spectral representation of stochastic differential equations (SDEs).

\subsection{Interpretable deep kernel}\label{sec:dkgp}

\begin{figure*}[h!]
    \centering
    \renewcommand{\tabcolsep}{-2.0mm}
    \begin{tabular}{p{0mm}*{1}{c}}
        & \includegraphics[width=1.8\columnwidth]{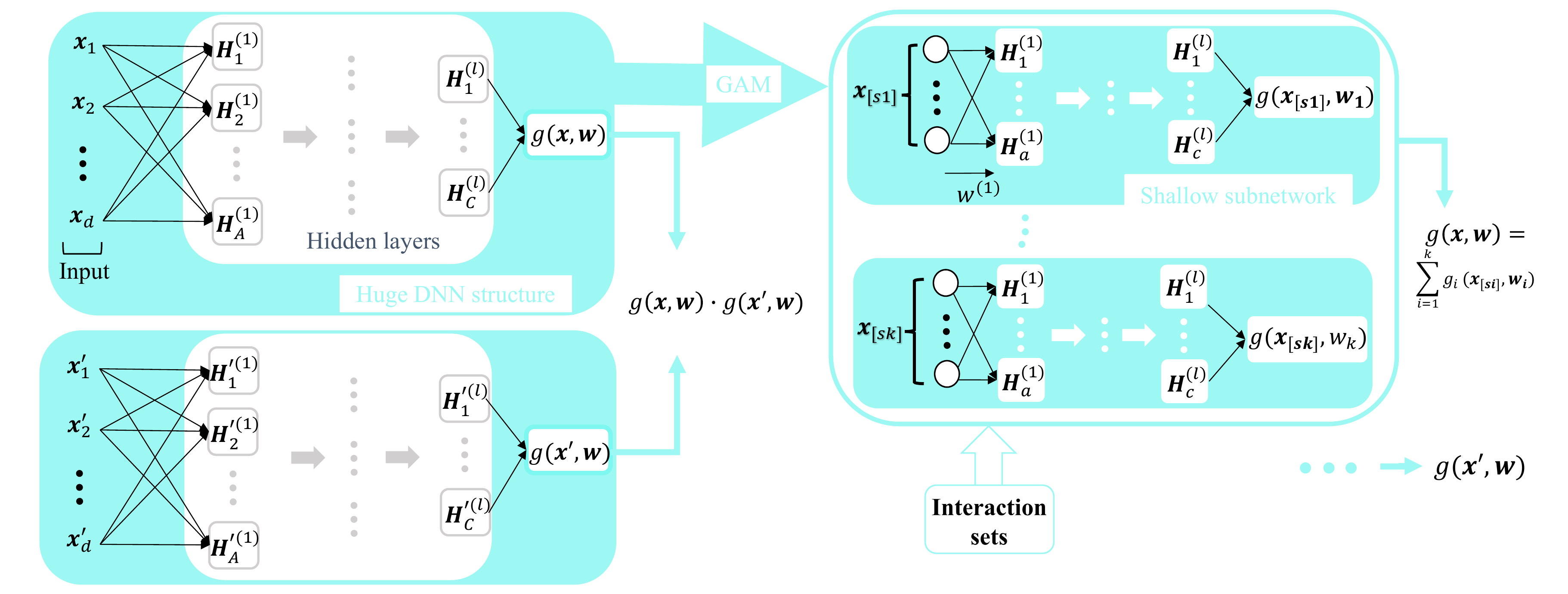}
    \end{tabular}
    \caption{The covariance structure of the optimal DKGP \cite{pmlr-v124-dai20a}, where a multilayer fully connected feed-forward NN is applied as the universal approximator of the underlying function $f(\vx)$. }
    \label{fig:deep}
\end{figure*}

Neal\cite{neal2012bayesian} proved that a Bayesian neural network with infinitely many hidden neurons converges to a GP. In practice, GPs with popular kernels are mostly used as simple nonlinear interpolation models. Deep neural
networks (DNNs) are demonstrated in their competent learning and representation in many application domains, including
computer vision \cite{DBLP:conf/nips/KrizhevskySH12}, speech recognition \cite{hinton2012deep}, language processing \cite{collobert2008unified}, and recommendation systems \cite{wang2015collaborative}. The most interesting DNN capability is feature discovery and representation. However, DNNs have a well-known interpretation imperfection in that the mechanism of model learning and inference is a black box, which heavily depends on hyper-parameter tuning techniques. Therefore, deep kernel GP (DKGP) \cite{wilson2016deep,al2017learning,wilson2016stochastic} combines the nonparametric flexibility of kernel methods with the inductive biases of deep learning architectures, which presents benefits in both expressive power and interpretability. As a result, the DKGP can draw their strengths to learn a model for complicated wireless mechanisms, such as 5G and vehicle-to-everything (V2X) channel impulse responses, multipath radio signal propagation, radio feature maps (such as the signal quality, uplink/downlink traffic, wireless resource demand/supply) over time and space, and indoor pedestrian motion, etc.

For DKGP, a typical framework extracts features from DNNs and then treats the features as inputs of multiple GPs \cite{wilson2016deep,wilson2016stochastic}. The model comes from linearly mixing these GPs and jointly optimizing hyper-parameters through a marginal likelihood objective. The understanding of this kind of deep kernel is straightforward and can actually be seen as the GP using complicated feature engineering or transformation before learning. The popular structure of the deep kernel can be written as
\begin{align}\label{eq:deepkernel}
k_{\deep}(\vx, \vx')\rightarrow\sum_{i=1}^{Q}{k_{i}\left(g_{\nn}(\vx, \vw_\nn), g_{\nn}(\vx', \vw'_\nn)\right)},
\end{align}
where $g_{\nn}(\vx, \vw)$ denotes a nonlinear feature mapping given by DNN with weights $\vw$.
Note that the kernel $k_{i}$ used in Eq. (\ref{eq:deepkernel}) can be arbitrary. Similar to the DNN, the chain rule is also applicable for deep kernel learning. According to the chain rule, the derivatives of the NLML with respect to the deep kernel hyper-parameters are given as follows:
\begin{subequations}
    \begin{align}
    \frac{\partial{\loss}}{\partial{\Theta}}&=\frac{\partial{\loss}}{\partial{K_{\deep}}} \frac{\partial{K_{\deep}}}{\partial{\Theta}},\\
    \frac{\partial{\loss}}{\partial{\vw_{\nn}}}&=\frac{\partial{\loss}}{\partial{K_{\deep}}} \frac{\partial{K_{\deep}}}{\partial{g_{\nn}(\vx, \vw_\nn)}} \frac{\partial{g_{\nn}(\vx, \vw_\nn)}}{\partial{\vw_\nn}},
    \end{align}
\end{subequations}
where the derivative of NLML with respect to the covariance matrix is $\frac{\partial{\loss}}{\partial{K_{\deep}}}=\frac{1}{2}\left(K^{-1}_{\deep}\vy\vy^{\top}K^{-1}_{\deep}-K^{-1}_{\deep}\right)$.

Another deep kernel using the finite rank Mercer kernel function with orthogonal embeddings on the last layer has a better learning efficiency and expressiveness \cite{dasguptafinite}. However, incorporating DNN into GP leads to poor interpretability due to DNN's blackbox. To enrich the interpretability of deep kernels, the second class of deep kernels was proposed to reveal the learning dynamics of the DNN by building connections between the GP and DNN. Furthermore, considerable focus has been paid on interaction detection in DKGP to enhance its interpretability. 
Interestingly, a recently proposed novel optimal DKGP (see Fig. \ref{fig:deep}) \cite{pmlr-v124-dai20a} demonstrates better model interpretability. The resulting kernel has a non-stationary dot product structure with minimized test mean squared error, shallow DNN subnetworks with feature interaction detection, much reduced hyper-parameter space, and good interpretability.

\subsection{Collective intelligence using multi-task kernel}\label{sec:mtgp}

In wireless communication systems, adjacent devices are not independent and must be correlated because there are shared patterns and environmental factors between them. For example, connected smartphones, robots, drones, vehicles, intelligent home systems, and NB-IoT sensors in the same wireless network may have dependent behaviors or trends impacted by the status of the wireless network. Hence, a joint learning model can make full use of data collected from adjacent devices to achieve collective intelligence. Knowledge obtained from different edge devices can be transferred to augment the overall prediction performance and system understanding. Therefore, a paradigm of multi-task learning can empower such collaboration in wireless communication.

The extension of GPs to multiple sources of data is known as a multi-task or multioutput Gaussian process (MTGP or MOGP). MTGP accounts for the statistical dependence across different sources of data (or tasks) \cite{bonilla2008multi,Ruan2017}. Given $m$ tasks, the aim of the MTGP model is to jointly learn $m$ underlying functions $\vf_{l}(X)=[f^{(1)}_{l}(\vx^{(1)}),...,f^{(m)}_{l}(\vx^{(m)})]^{\top}$ and estimate their function values $\vy=[y^{(1)},...,y^{(m)}]^{\top}$, where $X=[\vx^{(1)},...,\vx^{(m)}]^{\top}$, $\vy\sim\N(\vf_{l}(X), \vepsilon{I})$, and $\vepsilon=[\epsilon^{(1)}, ..., \epsilon^{(m)}]$ represents the noise variances of the $m$ tasks. However, similar to a single GP, underlying functions $\vf_{l}(X)$ still have a Gaussian distribution with $\vf_{l}(X)\sim\gp(\vzero, K_{\mtgp}(\vx^{(m)}, \vx^{(m')}))$. Usually, MTGP with $m$ tasks has a computational complexity of $\complexity(m^{3}n^{3})$ when the size of the training data in each task is $n$.
If a point $\vx$ comes from a task $m$ and $\vx'$ comes from another task $m'$, then their covariance is
\begin{align}
k_{\mtgp}(\vx, \vx) = k^{\mm}(\vx-\vx').
\end{align}
For MTGP, a crucial point is how to jointly encode the shared structure and difference between tasks in the kernel \cite{Yu2018}. Kernel design should consider both the cross-covariance between tasks and auto-covariance within each task. Early MTGP approaches mainly focus on linear combinations and convolution of independent single-source GPs, which correspond to the linear model of coregionalization (LMC) framework \cite{bonilla2008multi,Wilson2011,Ulrich2015} and convolved GP \cite{alvarez2012kernels,Alvarez2011}, respectively. Many improvements and applications of MTGPs have been introduced in previous works, such as \cite{bonilla2008multi,Wilson2011,Alvarez2011,Parra2017}.
One method for promoting the representation ability of MTGP model is via using the SM kernels. First, the SM-LMC kernel\cite{Wilson2011} models the covariance of a single task with an SM kernel, linearly combines these single tasks with LMC and provides an interpretation of the Gaussian process regression network (GPRN) from the perspective of a neural network with
\begin{align}
{K}_{\gprn}^{\mm}=\sum_{i=1}^{Q}{B_{i}}\otimes{k_{\sm,i}},
\end{align}
where $k_{\sm,i}$ is a covariance structure shared by tasks and $B_{i}$ encodes the cross-covariance between tasks.
Then, the cross-spectral mixture (CSM) kernel \cite{Ulrich2015} additionally introduced a phase factor into $B_{i}$ to encode amplitude and phase for cross-covariance with
\begin{align}
{K}_\csm^{\mm}=\sum_{i=1}^{Q}{B_{i}}\otimes{k_{\textit{SG},i}}(\tau;{\Theta}_{i}),
\end{align}
where ${k_{\textit{SG}i}}(\tau;{\Theta}_{i})$ is the phasor notation of a spectral Gaussian kernel. The multioutput spectral mixture kernel (MOSM) \cite{Parra2017} further represents both time and phase delay in cross covariance between tasks by using complex-valued matrix decomposition.

However, MOSM has a compatibility drawback in that it cannot reduce to the SM kernel when only one task is available. Therefore, a multioutput convolution
spectral mixture (MOCSM) kernel \cite{chen2019multioutput} was proposed to enjoy the compatibility property perfectly through cross convolution of time and phase delayed SM components. Another important extension of MTGP is multi-task generalized
convolution SM (MT-GCSM) kernel \cite{chen2020generalized}, which models nonlinear task correlations and dependence between arbitrary components and provides a framework for heterogeneous tasks with different levels of complexity.
The later convolved GP is more flexible and expressive because it allows each task to have its own kernel and complexity.

\section{Scalable distributed Gaussian process}\label{sec:dis-gp}

The distributed Gaussian process (DGP) in wireless communication involves learning on distributed edge devices. The use of DGP can avoid frequent interactions with a central server and allow each edge device to possess a local learning model. For delay-sensitive applications such as self-driving vehicles and unmanned aircraft, a local learning model can rapidly respond to a local request in a timely manner. Particularly, the DGP can save the overall time cost of the wireless communication when the central server is not available or network congestion occurs. Therefore, DGP can be seen as a form of on-device intelligence, which addresses the major concerns of scalable computation and privacy protection in wireless communication.

In this section, we introduce the framework of DGP, which has shown significant advantages in computational efficiency \cite{DBLP:conf/icc/XuYXLC19,DBLP:journals/jstsp/YinG17,DBLP:journals/pami/TavassolipourMS20,DBLP:conf/icml/PengZZQ17}. There are many reasons for the selection of DGP, such as scaling ordinary GP to large datasets, applying ordinary GP to distributed edge dataset, preventing access to privacy-sensitive data
and making full use of multicore high-performance computers (HPCs). 
In general, DGP splits big data into multiple ($M$) smaller pieces computed on local computing nodes to speed up the inference of the whole model \cite{DBLP:conf/nips/GalWR14}, which refrains from centrally collecting and storing massive data. 
The initial aim of DGP is to make GP scalable to big data. However, with the development of multicore computing architecture and edge computing in IoT networks, DGP is gradually receiving attention from research and industrial applications because it provides a more practical machine learning framework than the existing GPs. Some representative DGP works have been published recently \cite{DBLP:conf/icc/XuYXLC19,DBLP:journals/jstsp/YinG17,DBLP:journals/pami/TavassolipourMS20,DBLP:conf/icml/PengZZQ17,DBLP:conf/nips/GalWR14,ng2014hierarchical}. By using the map-reduce framework and decoupling the data conditioned on the inducing points, a distributed variational inference for GP and latent variable models (LVMs) was proposed \cite{DBLP:conf/nips/GalWR14} . The distributed variational inference for GP still has the limitation of scalable inference when the data size is $n\geq{10^{7}}$.
Another DGP is based on the mixture-of-experts (MoE) model \cite{DBLP:conf/icml/NguyenB14}. The MoE model weights the predictions of all local expert models (node) to give the final prediction. For MoE, a confusion is how to specify the number of experts and weight of each expert. Compared with MoE, product-of-GP-experts models (PoEs) \cite{ng2014hierarchical} that multiply predictions of independent GP experts can avoid assigning weight to experts but are inevitably overconfident. The marginal likelihood $p(\vy|X, \Theta)$ of PoEs is written as follows:
\begin{align}
p(\vy|X, \Theta) \approx\prod_{i=1}^{M}p^{(i)}(\vy^{(i)}|X^{(i)}, \Theta),
\end{align}
where $M$ is the number of GP experts and $p^{(i)}(\vy^{(i)}|X^{(i)}, \Theta)$ is the marginal likelihood of the $i$-th GP expert using the $i$-th partition $\{X^{(i)}, \vy^{(i)}\}$ of dataset $\{X, \vy\}$. Additionally, the predictive probability of PoEs is the product of all predictive probabilities of independent GP experts,
\begin{align}
p(f_{*}|\vx_{*}, \vy, X) \approx\prod_{i=1}^{M}p^{(i)}(f_{*}|\vx_{*}, \vy^{(i)}, X^{(i)}).
\end{align}

Similarly, the Bayesian committee machine (BCM) \cite{DBLP:journals/neco/Tresp00} combines independent estimators trained on different datasets by using Bayes' rule. BCM has a better interpretation due to considering the GP prior $p(f_{*})$. 
Furthermore, robust BCM \cite{DBLP:conf/icml/DeisenrothN15} generalized the original BCM and PoE-GP by incorporating a GP prior and the importance of GP experts. In order to achieve a much better approximation of a full GP, other improved DGP works include: (1) asynchronously distributed variational GP \cite{DBLP:conf/icml/PengZZQ17}
that uses weight-space augmentation to scale up GPs to billions of samples; (2) generalized robust BCM \cite{DBLP:conf/icml/LiuCWO18} that gains a consistent aggregated predictive distribution by randomly selecting a subset $\D^{(1)}$ as a global node for communicating with the remaining subsets; (3) nested kriging predictors that aggregates submodels based on subsets of observation points\cite{DBLP:journals/sac/RulliereDBC18}.

\section{An example of data-driven wireless communication using GP: the prediction of online 5G users}\label{sec:gp-example}

\begin{figure*}[h!]
    \renewcommand{\tabcolsep}{0.2mm}
    \def\figpred#1{\includegraphics[width=0.50\columnwidth]{fig-LSM-covSM-24h_date_number_value_cell_2_beam_27-Q6-#1}}
    \def\figcov#1{\includegraphics[width=0.50\columnwidth]{fig-LSM-covSM-24h_date_number_value_cell_2_beam_27-Q6-Cov-#1}}
    \def\figspec#1{\includegraphics[width=0.50\columnwidth]{fig-LSM-covSM-24h_date_number_value_cell_2_beam_27-Q6-Spec#1}}
    \begin{tabular}{p{0.5mm}*{4}{c}}
        & \figpred{} & \figpred{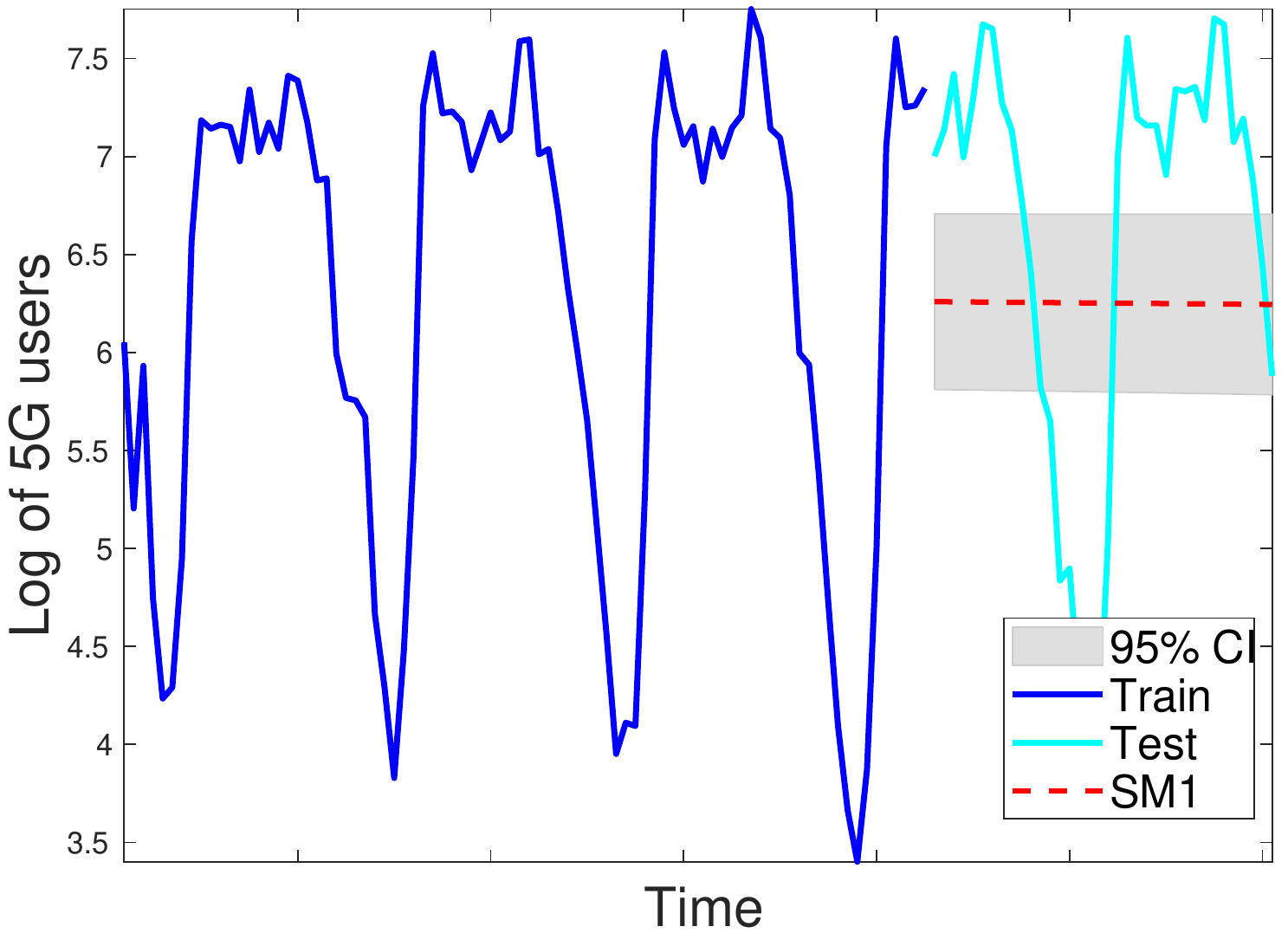} & \figpred{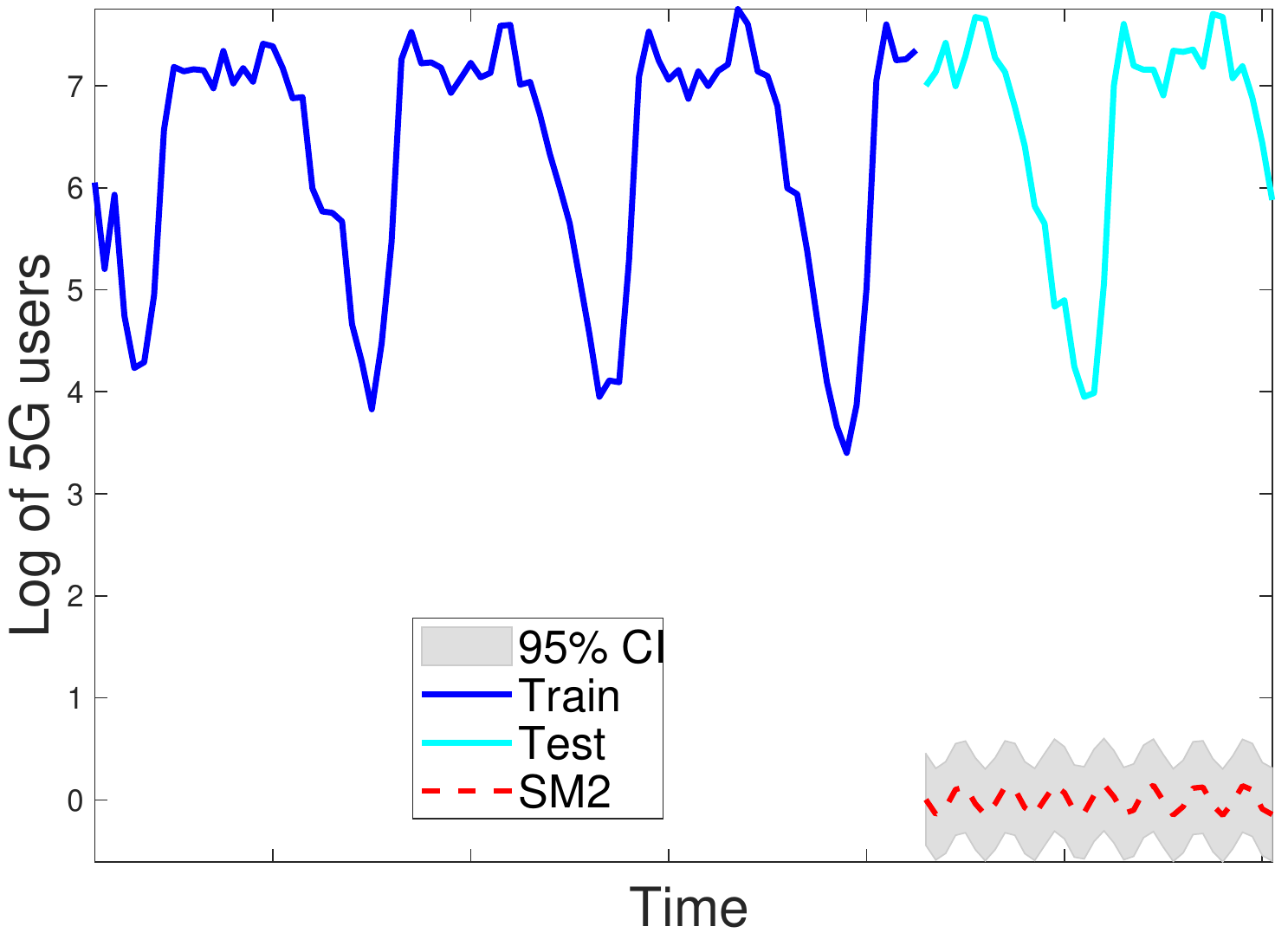} & \figpred{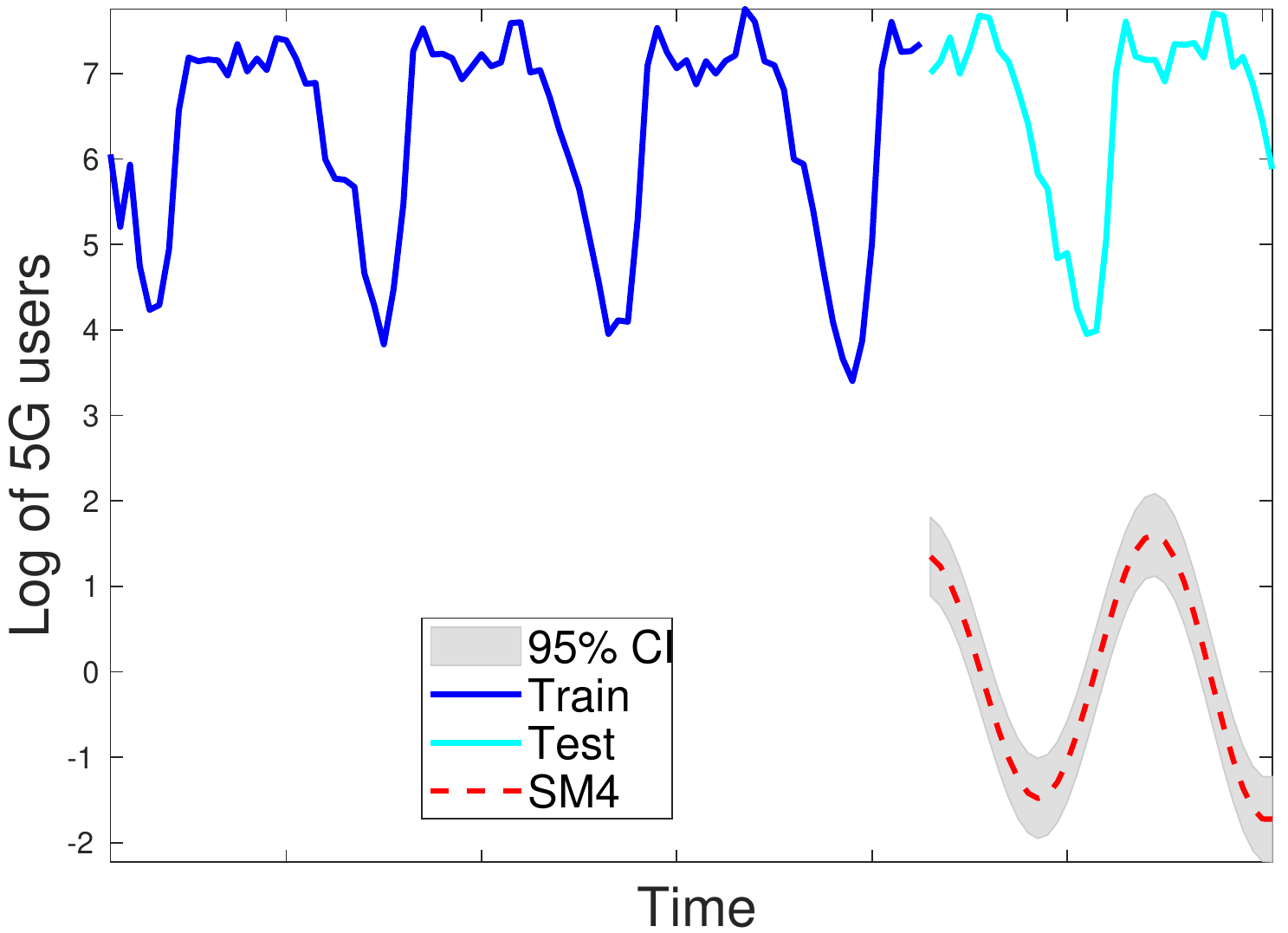} \\ 
        & {(a) $f\sim\gp(0, k_{\sm}(\tau))$} & {(b) $f_{1}\sim\gp(0, k_{\sm, 1}(\tau))$} & {(c) $f_{2}\sim\gp(0, k_{\sm, 2}(\tau))$} & {(d) $f_{4}\sim\gp(0, k_{\sm, 4}(\tau))$} \\
        & \figcov{} & \figcov{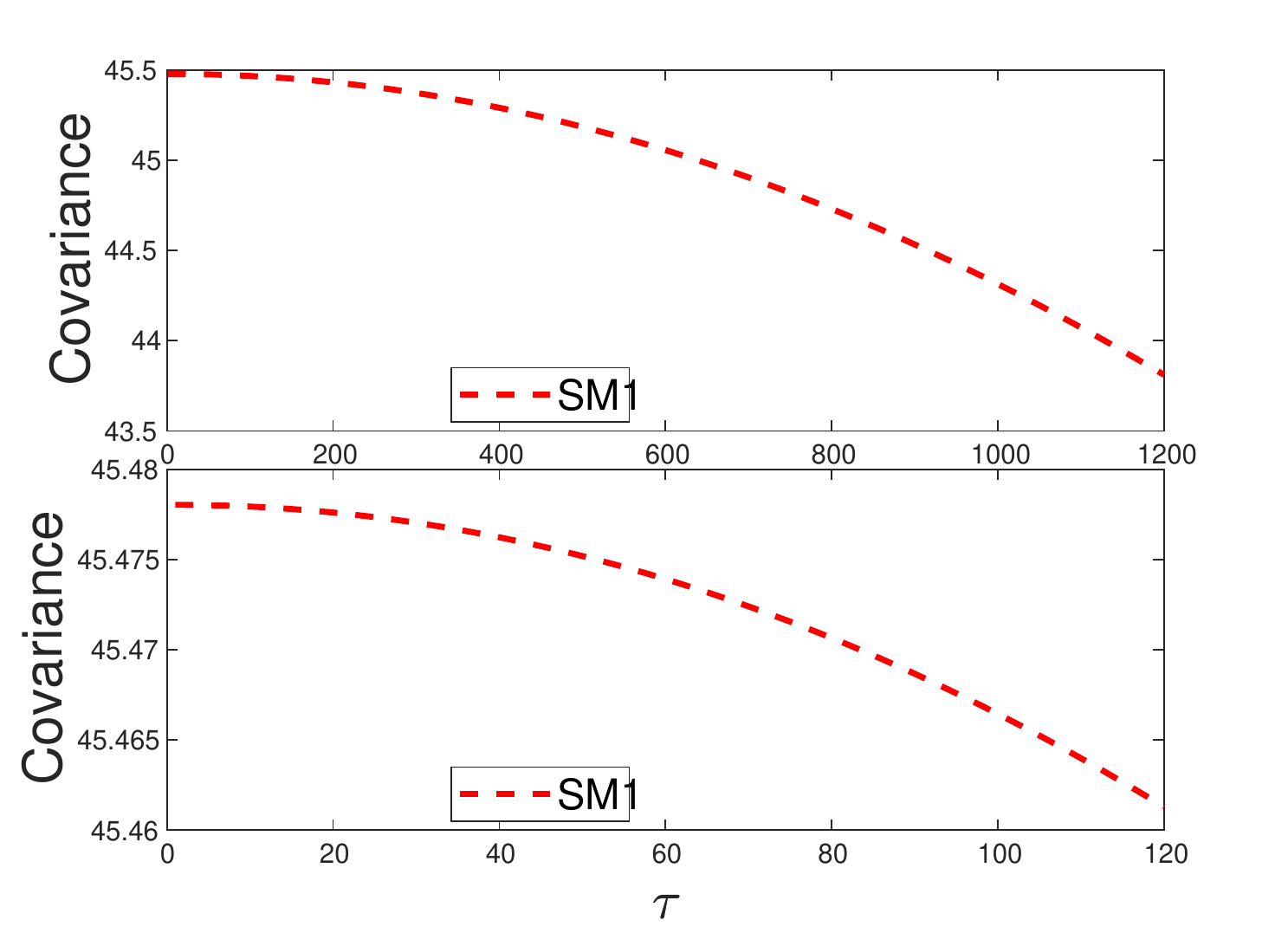} & \figcov{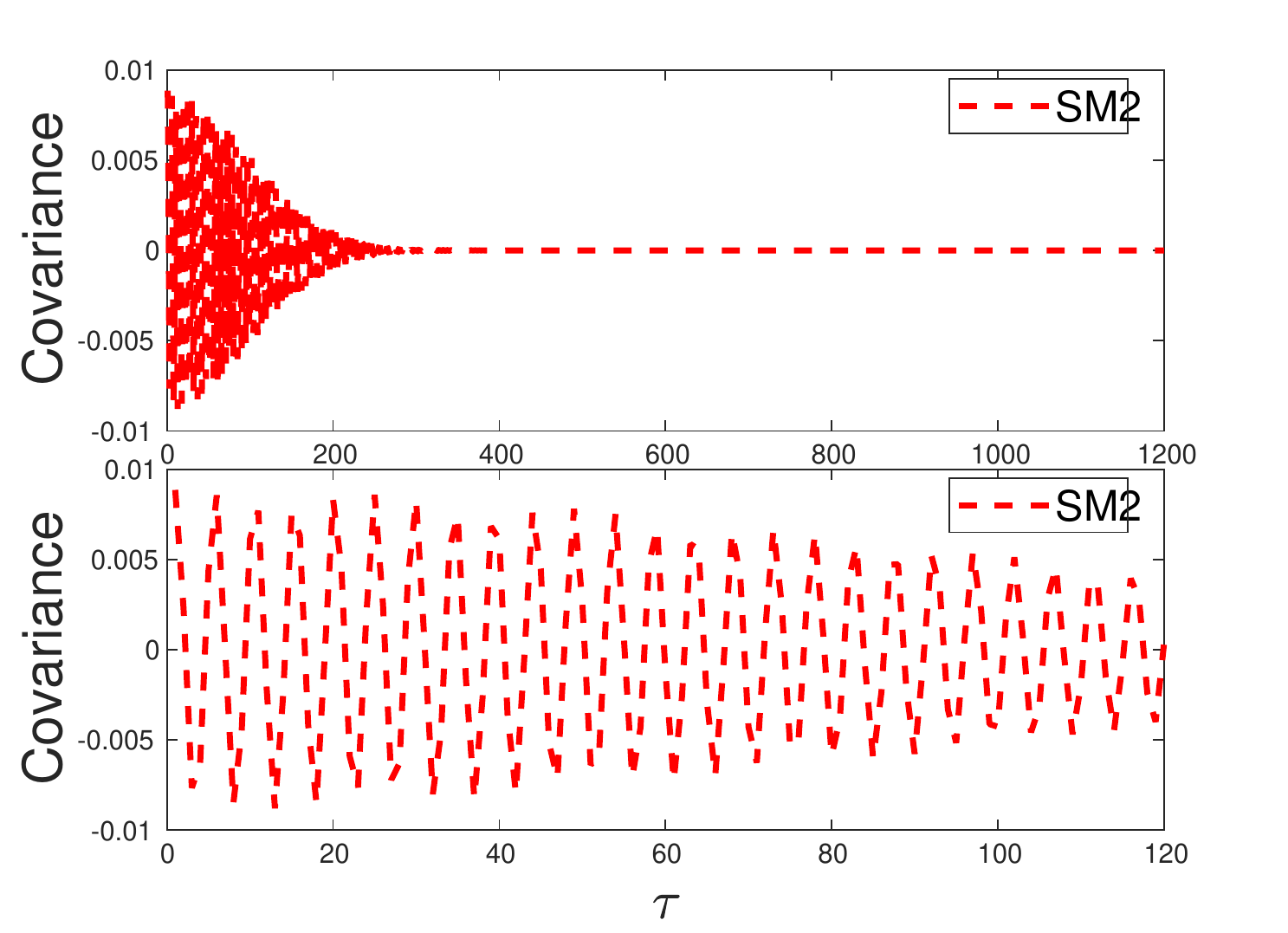} & \figcov{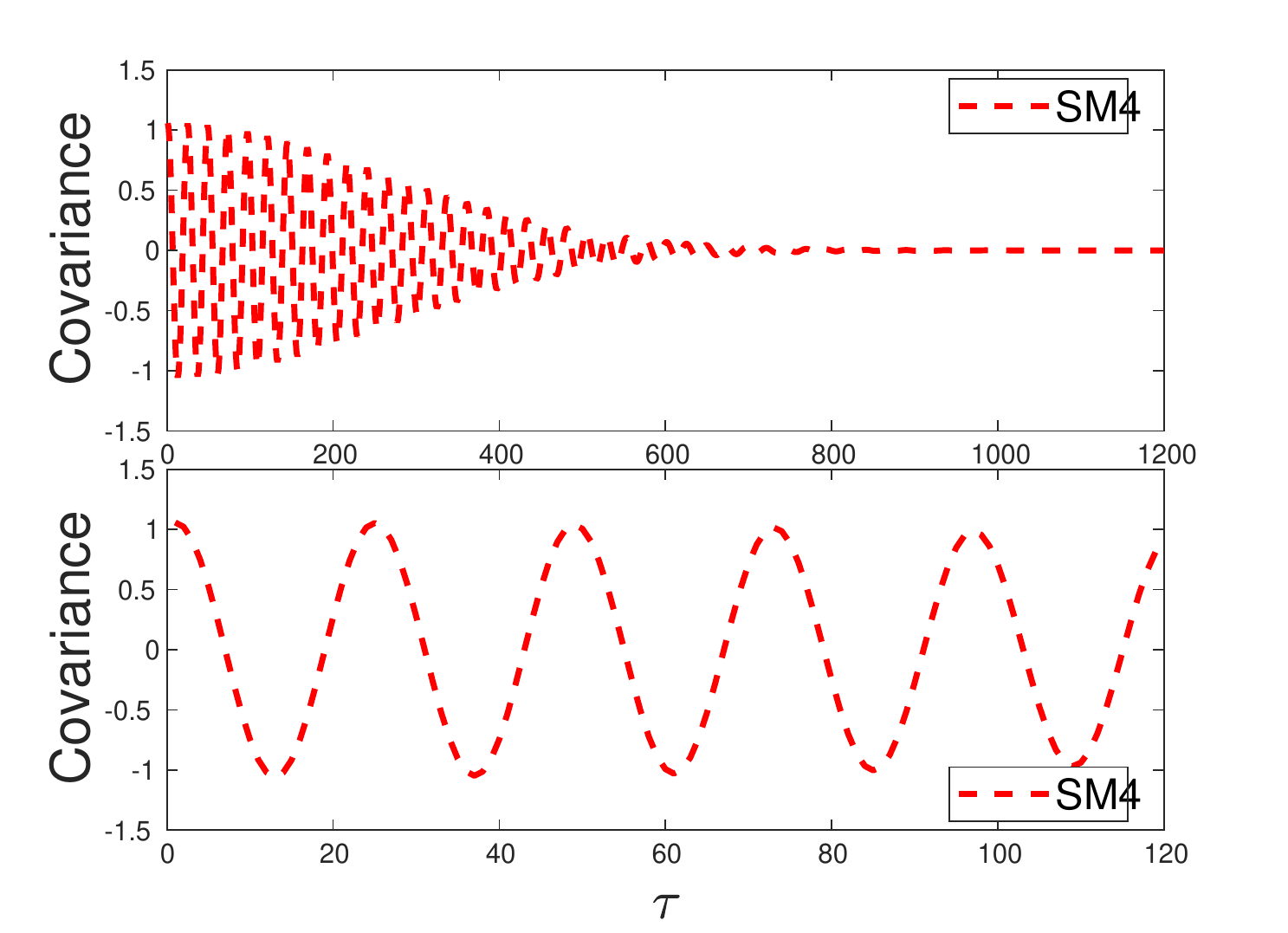} \\
        & {(e) ${k}_{\sm}(\tau)$} & {(f) ${k}_{\sm, 1}(\tau)$} & {(g) ${k}_{\sm, 2}(\tau)$} & {(h) ${k}_{\sm, 4}(\tau)$} \\
        & \figspec{} & \figspec{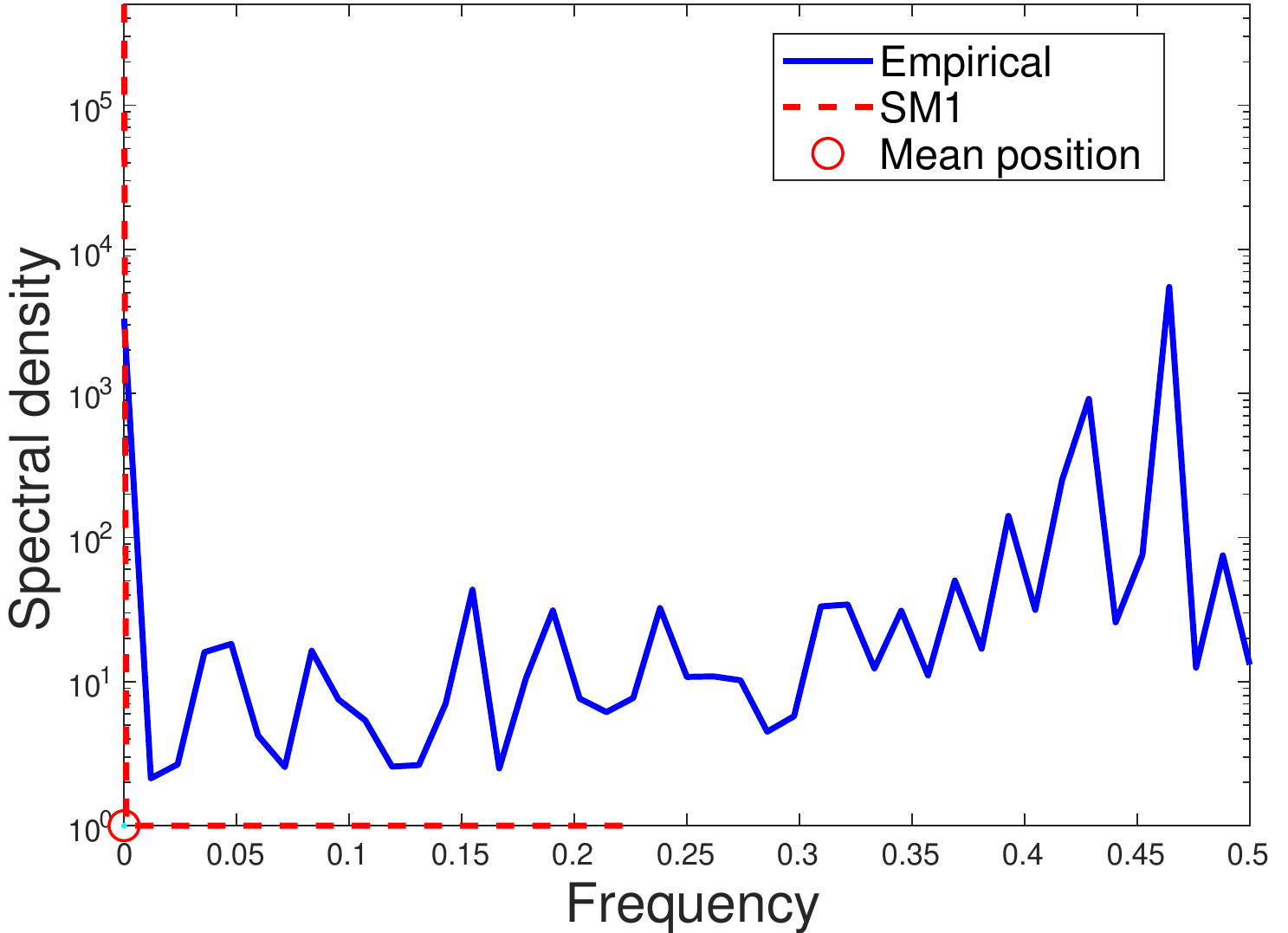} & \figspec{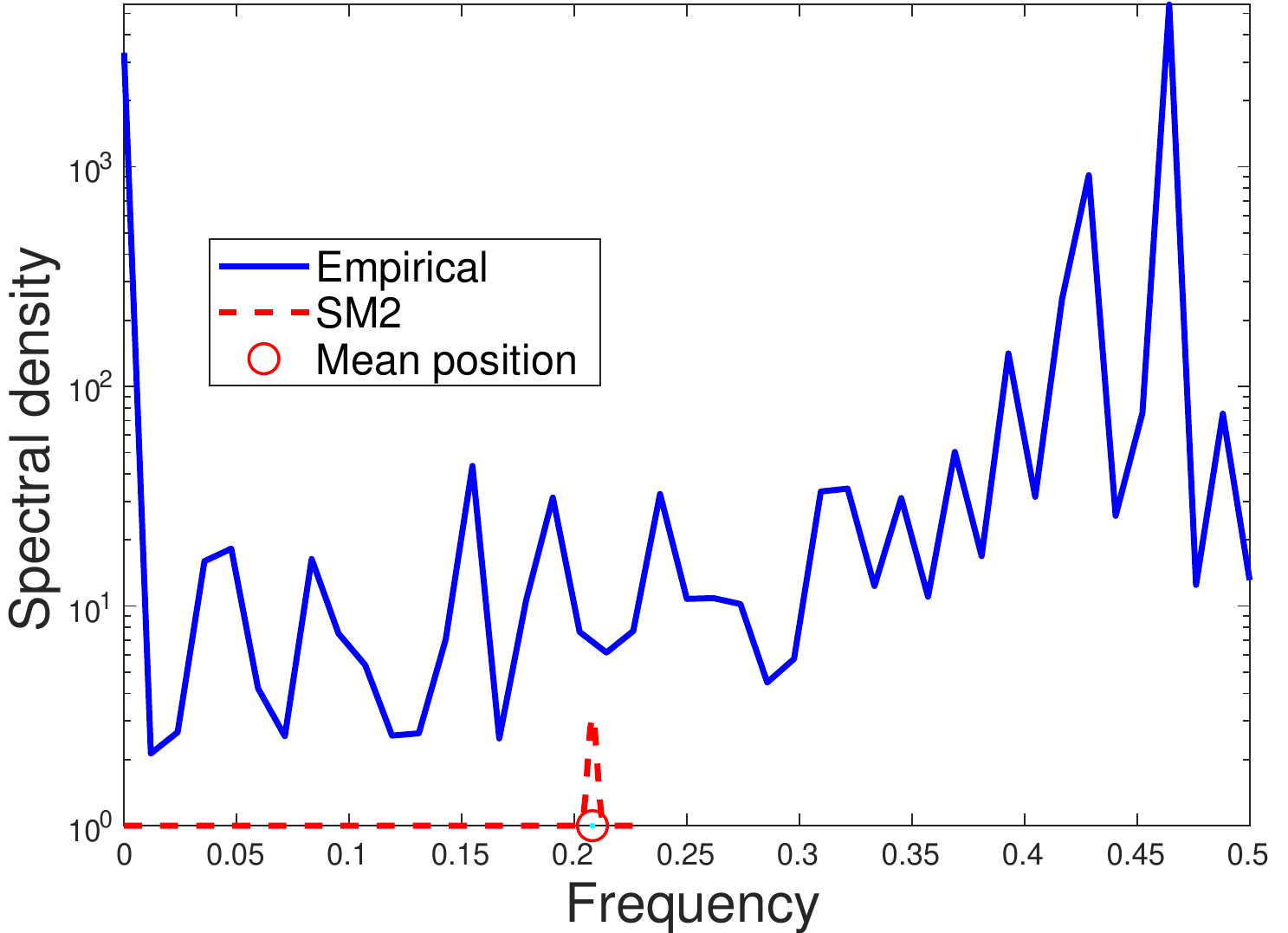} & \figspec{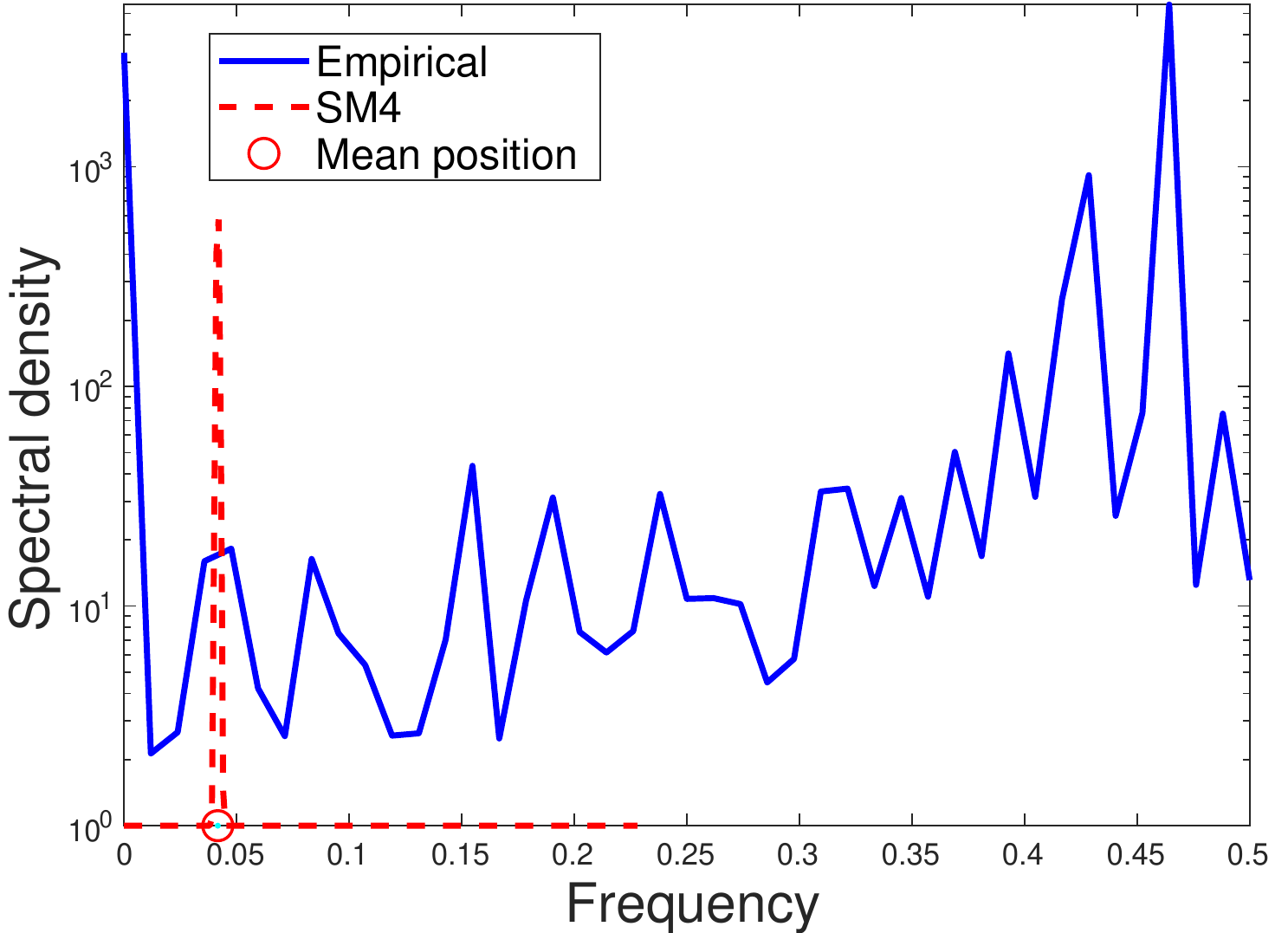} \\
        & {(i) $\freq{k}_{\sm}(s)$} & {(j) $\freq{k}_{\sm, 1}(s)$} & {(k) $\freq{k}_{\sm, 2}(s)$} & {(l) $\freq{k}_{\sm, 4}(s)$}
    \end{tabular}
    \centering
    \caption{Prediction of online 5G users using a GP model with the SM kernel $k_{\sm}$, where ${k}_{\sm}(\tau)=\sum_{i=1}^{6}{k}_{\sm, i}(\tau)$ and its spectral density $\freq{k}_{\sm}(s)=\sum_{i=1}^{6}\freq{k}_{\sm, i}(s)$. Subplots in the first row: the prediction of the GP model and its predictive trends captured by the 1st, 2nd, and 4th SM components. Subplots in the second row: the corresponding short and long range covariances of the GP model and its predictive trends. Subplots in the last row: the corresponding spectral densities of the SM kernel and its components.}
    \label{fig:gp-example}
\end{figure*}

In this section, we show an exemplary case by considering a real world wireless communication dataset regarding the number of online 5G users collected from a 5G base station.
This experiment can substantiate the features of data-driven wireless communication using GPs in terms of expressiveness, interpretability, and uncertainty modeling. The dataset was collected 
in a southern city in China in 2021.
As shown in Fig. \ref{fig:gp-example}, there are multiple patterns with different time scales in the varying of the number of online 5G users, such as long-term (subplot (b)), short-term (subplot (c)), and mid-term (subplot (d)) trends.
We split $70\%$ of the dataset as training (in blue, up subplots) data  and the rest $30\%$ as testing (in cyan, up subplots) data.
We set the SM kernel with $Q=6$ components and initialize the hyper-parameters by fitting a Gaussian mixture model on the empirical spectral densities (in blue, bottom subplots). 
Subplot (a) in Fig. \ref{fig:gp-example} indicates that the GP model (in dashed red) can extrapolate the patterns of testing well. The predictive trend (in dash red) is very close to the ground true trend (in cyan). Moreover, the predictive $95\%$ confidence interval (CI) (in grey shade) of the GP model shows uncertainty bounds of the prediction and completely covers the ground true.

The interpretability of the GP model can be sniffed from the structure of spectral density (in dash red, bottom subplots). As shown in subplot (i) Fig. \ref{fig:gp-example}, there are six significant peaks captured by the GP model, which denotes different patterns with different periods. The peaks located at low frequencies reveal long-term trends with period $\mu_i^{-1}$ . On the contrary, the peaks located at high frequencies describe short-term trends. These trends with different time scales connect to the periodicity of human activities. The covariance of long-term trend in subplot (f) decays much slower than mid-term trend in subplot (g) and short-term trend in subplot (h). The amplitudes of spectral peaks determine the  scales of their corresponding function values. Specifically, we exhibit three patterns (the 1st, 2nd, and 4th) learned by the GP model and their corresponding spectral densities. For instance, the 4th pattern in subplot (d) demonstrates a mid-term evolution of online 5G users, which has a spectral density peak located at $\mu_{4}=0.04$. The sum of all these 6 spectral peaks constitutes the learned spectral structure of the GP model, namely, the Fourier transform of the covariance of the GP model. 
In this context, all learned patterns have clear physical interpretations. 
However, for neural networks we cannot get such interpretations with insights for each neuron.  The expressiveness of the GP model is guaranteed due to the theories, namely, the generalization of Fourier transform and the approximation of Gaussian mixture on spectral distribution. In particular, the expressiveness of the GP model can be further improved by giving more components when learning from high complicated wireless task. By using the distributed framework shown in Section \ref{sec:dis-gp}, the scalability of the GP model is easily achieved for large wireless dataset.

\section{GP based wireless applications}\label{sec:wireless-gp}

In this section, to show the widespread applications of GPs in wireless communication, we further review and discuss three typical applications: wireless traffic prediction, localization, and trajectory planning.

\subsection{Wireless traffic prediction}\label{sec:gp-traffic}

Wireless traffic prediction has been a long-standing demand for wireless network planning and management. Highly accurate wireless traffic prediction can reduce the uncertainty of network load and reflect the traffic behavior in wireless network, which greatly matches the benefits of GP in terms of uncertainty and interpretability.
There are many traffic related issues in wireless communication suitable for GP model, such as wireless traffic analysis \cite{DBLP:conf/icc/XuYXLC19,DBLP:journals/jsac/XuYXLC19}, cellular traffic load prediction \cite{wang2020cellular}, traffic load balancing for multimedia multipath systems \cite{he2019calibrating}, channel prediction for communication-relay UAV \cite{ladosz2019gaussian}, stochastic link modeling of static wireless sensor networks \cite{shahanaghi2019stochastic}. We survey some representative examples as follows:

\begin{itemize}
    \item {\bf Wireless traffic prediction}. In \cite{DBLP:journals/jsac/XuYXLC19,DBLP:conf/icc/XuYXLC19}, a GP model with the alternating direction method of multipliers (ADMM) for distributed hyper-parameter optimization was proposed to predict 4G wireless traffic, which shows better performance than a DNN model, such as long short-term memory (LSTM).

    \item {\bf Cellular traffic load prediction}. In \cite{wang2020cellular}, a scheme combining GP and LSTM was proposed to generate accurate cellular traffic load prediction, which is important for efficient and automatic network planning and management. Compared with benchmark schemes, the proposed scheme achieves state-of-the-art performance.
        
    \item  {\bf Traffic load balancing for multi-media multi-path systems}.
    In \cite{kim2019gaussian}, an adaptive load balancing algorithm using an online GP was proposed to estimate the path status and allocate traffic load to each path properly in multimedia multipath systems. The proposed GP based algorithm is helpful for offering higher reliability and stability utilizing a variety of communication media and paths.        
\end{itemize}

\subsection{Wireless localization}\label{sec:gp-locate}

Another representative wireless scenario using GPs is wireless localization.
Wireless localization has become a cornerstone of modern life due to the increasing demand of location-based application, e.g., shopping and industry activities. 
GPs have been successfully applied for indoor wireless localization due to its uncertainty modeling and nonlinear regression abilities, such as  wireless tracking \cite{DBLP:journals/jstsp/YinG17},
online radio map update \cite{xu2019online}, and calibrating multichannel RSS observations for localization \cite{he2019calibrating}. We review various indoor localization frameworks using GPs as follows:

\begin{itemize}
    \item {\bf Wireless target tracking}. In \cite{DBLP:journals/jstsp/YinG17}, a framework of distributed recursive GP was proposed to build multiple local received signal strength (RSS) maps, which has reduced computational complexity on big data generated from large-scale sensor networks. Then, a global map is constructed from the fusion of all the local RSS maps. The proposed framework shows excellent positioning accuracy in both static fingerprinting and mobile target tracking scenarios.

    \item {\bf Online radio map update}. In \cite{xu2019online}, a novel scheme combining crowdsourcing and GP regression can adapt radio maps to environmental dynamics in an online fashion, which recursively fuses crowdsourced fingerprints with an existing offline radio map. The scheme has particular advantages in efficiency and scalability.
    
    \item {\bf Calibrating multichannel RSS observations for localization}. In \cite{he2019calibrating}, a GP model was proposed to compensate for frequency-dependent shadowing effects and multipaths in received signal strength (RSS) observations. By applying the GP model, multichannel RSS observations can be more effectively combined for localization over a large space.
\end{itemize}

\subsection{Wireless trajectory planning}

UAV has been an emerging technology strongly correlated with wireless communication. 
Due to the complex communication environment and irregular time delay in wireless network, UAV trajectory planning is rapidly gaining attentions in wireless communication.
To meet the requirement of expressiveness appearing in wireless network, 
popular RL and DNN methods \cite{hu2020cooperative} have recently considered as panaceas for wireless trajectory planning.

In addition,  the GP-based approaches with uncertainty representation and sample efficiency are also impressive for wireless trajectory planning because of their unique features, namely, model expressiveness, uncertainty representation, and sample efficiency.
We review recent advances of wireless trajectory planning using GPs as follows:

\begin{itemize}
    \item {\bf GP-based runtime planning, learning, and recovery for safe UAV operations}. In \cite{9341641}, a recursive dynamic GP regression-based framework with fast online planning, learning, and recovery approach was proposed for safe UAV operations under unknown runtime disturbances. The framework can estimate the behavior of the UAV system and provide safe plans at runtime under unseen disturbances.
    
    \item {\bf Networked operation of a UAV using GPs}. In \cite{jang2019networked}, a novel GP model predictive control (MPC) scheme for path planning was introduced to  deal with the time-varying network delay, non-linearity, and time-sensitive characteristics of multirotor-type UAVs. The scheme increased the accuracy of path planning and state estimation for multirotor-type UAVs. 
    
    \item {\bf Obstacle-aware informative path planning using GPs
        for UAV-based target search}. In \cite{meera2019obstacle},  an algorithm leverages a layered planning strategy using a GP-based model of target occupancy to generate informative paths in continuous 3D space. The algorithm can achieve a balance between information gain, field coverage, sensor performance, and collision avoidance for efficient target detection.
\end{itemize}

\section{GP for future wireless communication}\label{sec:future}

From the motivations of using GP for wireless communication, we note that there are many emerging difficulties.
We outline a few challenging open issues of GP models for future data-driven wireless communication.

\begin{itemize}
    \item {\bf Ultra large-scale distributed GPs for dense and decentralized wireless communication systems}. In future data-driven wireless communication, the widely existing sensors gather considerable data at all times, which leads to large considerable data transmission and storage. An effective and pragmatic solution reducing the cost of data transmission and storage is to perform ultra large-scale distributed machine learning.    
    Even though the scalability of GP is available currently. However, ultra large-scale distributed GP is still an open research issue.
    \item {\bf GPs for 
        multimodal data in wireless communication systems}.
    Currently, the GP model can only learn from structured data generated from wireless communication systems. There are also multimodal data collected from different types of sensors, such as numerical raw data from smartphones, ultrasound data from UAV ultrasonic sensors, images and videos from surveillance cameras, and natural language from speech sensors. Particularly, the signaling and data transmitted via interfaces of both LTE/5G wireless and core networks, are always nonstructured.   
    Therefore, learning from nonstructured and multimodal data in wireless communication systems is another challenge for GP model.
    
    \item {\bf Highly interpretable GPs with deep structure in wireless communication systems}. The deep kernel of a GP has difficulties in that it increases the flexibility of the GP model as well as the difficulty of model interpretation. From both the theorems of stationary and non-stationary kernels, the mathematical definition of deep kernels in the frequency domain remains unclear. Similar to DNN, sacrificing interpretability in data-driven wireless communication is usually the compromise option between learning and understanding the network, which is less tolerable for high complexity state prediction. Hence, pursuing a highly interpretable GP with deep structures will be a critical open issue in future data-driven wireless communication.
\end{itemize}

\section{Conclusion}\label{sec:con}
In this paper, we comprehensively review data-driven wireless communication using GPs in terms of motivation, definition and construction of a GP model, GP expressiveness using different kernels, and distributed GP scalability. A GP with a Bayesian nature can model a large class of wireless communication systems through the designation of its covariance function. By using a distributed approach, GP models are capable of performing scalable inference on big data in a wireless network.

Data-driven wireless communication systems using GPs can achieve desired properties, expressiveness, scalability, interpretability, and uncertainty modeling. These characteristics become crucial for models in wireless communication due to the collected rich data and the modeling complexity in wireless networks.
In particular, interpretability and uncertainty modeling are inherent advantages of GPs due to their mathematical definition. From existing applications of the GP models in wireless communication, we present that the GP models can cover the aforementioned properties of data-driven wireless communication very well, which has been successfully proven to be valuable.

\bibliographystyle{IEEEbib}
\bibliography{GP-com}
\end{document}